%% file: main.tex
\documentclass[11pt]{article}
\usepackage[a4paper,margin=2.6cm]{geometry}
\usepackage[T1]{fontenc}
\usepackage{lmodern}
\usepackage[numbers,sort&compress]{natbib}
\usepackage{amsmath,amssymb}
\usepackage{booktabs,multirow,array}
\usepackage{graphicx}
\usepackage{tikz}
\usetikzlibrary{positioning,calc}
\usepackage{caption}
\usepackage[hidelinks]{hyperref}
\usepackage{xspace}
\input{tables/numbers.tex}


\newcommand{\tfm}{TFM\xspace}
\newcommand{\tfms}{TFMs\xspace}

\begin{document}

\title{Tabular Foundation Models vs Tuned Gradient Boosting\\
for Telecom Churn: A Profit- and Calibration-Aware Evaluation}
\author{Ismail Hossain Polas\\ Jahangirnagar University\\ \texttt{ismailhossainpolash@gmail.com}}
\date{}

\maketitle

\begin{abstract}
A churn model is useful through the customers it selects for a retention offer: the better model is the one whose
selection makes the campaign more profitable. Tabular foundation models (\tfms), pretrained networks that predict from
a table of labelled examples without being trained on it, were recently reported to decisively outperform tree
ensembles for churn. That comparison, however, used untuned gradient-boosted trees (GBDTs) and ranking metrics only. We
ask whether the advantage holds against tuned GBDTs when models are judged by expected campaign profit and by the
reliability of their probabilities. Four \tfms (TabPFN-2.5, TabPFN-3.5, TabPFN-3.5-Fast, TabICL v2) face LightGBM, XGBoost and
CatBoost tuned with Optuna on validation AUC, on five public telecom datasets, three from real operators, with \nSeeds{} repeated
splits each and \nPrimary{} pre-declared comparisons tested with a corrected $t$-test and Holm correction. TabPFN-3.5
earns significantly more expected maximum profit (EMP) than each tuned GBDT on two datasets, Cell2Cell
(\empPFNThreeCellTwoCellAbs\%) and Iranian (\empPFNThreeIranianAbs\%). On Orange it is ahead in every split
(\empPFNThreeOrangeAbs\%), but not significantly after correction, and no \tfm differs significantly from the GBDTs on the
two fictional datasets.
Across \nScenarios{} assumed combinations of customer value, incentive cost and acceptance rate, TabPFN-3.5 stays ahead on
all three real-operator datasets, though the size of its gain varies widely. Retuning the GBDTs on EMP moved no gap by
more than \maxEmpTuneShift{} per customer.
Ranking and probability reliability diverge: raw TabICL v2 underpredicts churn on Orange, and a threshold policy built on
its probabilities loses \regICLOrangeRaw\% of attainable profit until it is recalibrated. Because the splits resample the
same customers, the results describe these datasets, not telecom operators in general. The non-commercial licence of
the TabPFN weights, rather than scoring cost, restricts production use.
\end{abstract}

\section{Introduction}

Telecom operators try to keep customers who are about to leave by offering them an incentive. Each offer costs money,
and it pays off only when it reaches a customer who would otherwise have left and who accepts it. A churn model
therefore matters through the list of customers it selects. The better model is the one whose list makes the campaign
more profitable, which is not always the one with the higher classification score. Verbraken et al.\ made this precise
with the expected maximum profit (EMP) criterion \cite{verbraken2013emp}, and later studies found that selecting churn
models by AUC rather than by profit can reduce the expected profit of a retention campaign
\cite{verbeke2012newinsights,hoppner2020proftree,manzoor2025eprofits}.

A small example shows what EMP measures. Suppose a retained customer is worth 200 currency units to the operator (the
customer lifetime value, CLV), the incentive costs 10 and each contact costs 1. Contacting a churner who accepts the
offer gains $200-10-1=189$; contacting a churner who declines loses 1; contacting a loyal customer loses 11, because
that customer takes the incentive without needing it. EMP sorts customers by predicted risk, chooses the share to
contact that maximises expected profit, and averages over uncertainty about how many churners accept. It is expressed
per customer in the base, so a model with a higher EMP selects a more profitable contact list under these assumed
economics. These values are illustrative defaults, not measurements from any operator, and Section~\ref{sec:scn}
varies them.

Tabular foundation models (\tfms) offer a different way to build such a model. TabPFN
\cite{hollmann2025tabpfn,priorlabs2025tabpfn25,priorlabs2026tabpfn35} and TabICL \cite{qu2025tabicl,qu2026tabiclv2}
are pretrained once on large numbers of synthetic tables. To score a new customer, the model reads the operator's
labelled customers as context and predicts in a single forward pass, with no training on that data and no
hyperparameter search. On general benchmarks these models now match or beat tuned gradient-boosted decision trees
(GBDTs) \cite{erickson2025tabarena,qu2026tabiclv2,priorlabs2026tabpfn35}, the usual choice for tabular data. For churn,
Seyedzadeh and Karimi \cite{seyedzadeh2026churn} compared six \tfms with classical, deep and tree-ensemble models on nine
Kaggle datasets and concluded that \tfms ``collectively and decisively outperform'' them, with TabICL v2 ahead of
XGBoost by up to 9.23 PR-AUC points.

That study was the first to bring \tfms to churn, and three features of its design affect how its conclusion can be
read. First, its XGBoost and random forest baselines used library-default hyperparameters
\cite[Section~3.2]{seyedzadeh2026churn}, so part of the reported gap may come from untuned baselines rather than from the
model class. Second, models were scored by F1, ROC-AUC, PR-AUC and log loss, and the financial impact was estimated by
scaling the PR-AUC gap by a retention rate rather than by evaluating a campaign. Third, the study's own discussion
describes TabICL v2's lead as ``operational \ldots\ rather than as a statistically decisive metric lead''
\cite[Section~4.2]{seyedzadeh2026churn}. Each of these choices can enlarge the apparent advantage of \tfms. Together
they leave open whether an operator that already uses a well-tuned GBDT would gain anything by switching.

This paper asks that question directly: \emph{against tuned GBDTs, and judged by expected campaign profit, does a
\tfm select a more profitable set of customers, and can its probabilities be used as they are to decide whom to
contact?} We separate three parts of the answer, because a model can do well on one and badly on another: whether it
ranks churners well (ROC-AUC, PR-AUC), whether its ranking earns more campaign profit (EMP), and whether its predicted
probabilities are reliable enough to drive a fixed decision threshold (calibration and policy regret).

We compare four \tfms (TabPFN-2.5, TabPFN-3.5, TabPFN-3.5-Fast, TabICL v2) with Optuna-tuned LightGBM, XGBoost and
CatBoost on five public telecom datasets. Three come from real operators (Cell2Cell, Iranian, Orange); two are
fictional teaching datasets (IBM Telco, Maven), included because prior churn studies use them. Each dataset is split
\nSeeds{} times. The claims rest on \nPrimary{} comparisons fixed in advance, tested with a corrected resampled $t$-test and
corrected for multiple testing; every other analysis is labelled exploratory. The contributions are:
\begin{enumerate}
\item \textbf{A test of the reported \tfm advantage under tuned baselines and a profit criterion.} TabPFN-3.5 earns
  significantly more EMP than each of the three tuned GBDTs on Cell2Cell and Iranian. On Orange it is ahead in every
  split, but the difference is not significant after correction, and on Telco and Maven no \tfm differs significantly
  from the GBDTs. The advantage belongs to a model generation and a dataset, not to the model class
  (Section~\ref{sec:rq1}). Across \nScenarios{}
  combinations of customer value, incentive cost and acceptance rate, it stays ahead on all three real-operator
  datasets, but the size of the gain depends strongly on the assumed economics (Section~\ref{sec:scn}).
\item \textbf{Evidence that ranking and probability reliability must be checked separately.} Profit metrics change
  the winning model only when models are statistically tied (Section~\ref{sec:rq2}). A model that ranks well can
  still give unreliable probabilities: raw TabICL v2 underpredicts churn on Orange, and a fixed-threshold campaign
  built on its probabilities loses \regICLOrangeRaw\% of attainable profit until it is recalibrated
  (Section~\ref{sec:rq3}).
\item \textbf{Checks against potential methodological concerns, each with its scope stated.} A class-balanced context
  reduces rather than increases the \tfm advantage when the full training set fits (Section~\ref{sec:rq4}). Tuning
  the GBDTs on EMP instead of AUC, on every split of the five datasets, changes no gap by more than \maxEmpTuneShift{}
  profit units per customer. A larger tuning budget for LightGBM and CatBoost and a duplicate-aware split, each on five
  splits of three datasets, leave the conclusions unchanged (Section~\ref{sec:sens}).
\item \textbf{A bounded replication.} On the dataset behind the earlier benchmark's headline, tuning the baseline
  removes about a third of TabICL v2's PR-AUC lead, and the rest remains (Section~\ref{sec:rq5}).
\item \textbf{Deployment terms.} With a cached context, scoring cost is small; the licences of the TabPFN-3.5 and
  TabPFN-2.5 weights we used, which exclude production use, are the binding constraint (Section~\ref{sec:rq6}).
\end{enumerate}
The code, every per-split prediction, the selected hyperparameters of every tuned GBDT, and a generator and audit for
every number in this paper are released (Section~\ref{sec:availability}).

\section{Related Work}

\paragraph{Profit-aware churn evaluation.}
Verbraken et al.\ defined EMP as the expected best campaign profit when the probability that a contacted churner
accepts the offer is uncertain and Beta distributed \cite{verbraken2013emp}. Verbeke et al.\ used profit to compare
churn classifiers in telecom \cite{verbeke2012newinsights}, and H\"oppner et al.\ optimised EMP directly when growing
decision trees \cite{hoppner2020proftree}. Manzoor et al.\ proposed e-Profits, which replaces EMP's population-level
customer value with a per-customer value derived from Kaplan--Meier retention, and found that it reranks six classical
models on IBM Telco and Maven \cite{manzoor2025eprofits}. We take our evaluation criteria from this line of work. It
predates \tfms, so it does not ask whether they change which model a profit-aware operator should choose.

\paragraph{Tabular foundation models.}
TabPFN v2 \cite{hollmann2025tabpfn} established prior-fitted transformers for tables up to roughly 10k rows;
TabPFN-2.5 extended this to 50k rows \cite{priorlabs2025tabpfn25}, and TabPFN-3 and 3.5 to a million rows with a cached
context for fast scoring \cite{priorlabs2026tabpfn3,priorlabs2026tabpfn35}. TabICL separates column and row attention to
scale in-context learning \cite{qu2025tabicl}; TabICL v2 reports beating a tuned and ensembled RealTabPFN-2.5 on TabArena
and TALENT without tuning \cite{qu2026tabiclv2}. On TabArena, GBDTs remain strong on larger datasets while \tfms excel on
small ones \cite{erickson2025tabarena}, and TabPrep shows that feature engineering, absent from most benchmarks,
changes these comparisons \cite{tschalzev2026tabprep}. TabPFN's predictive probabilities are reported to be well
calibrated out of the box \cite{tanna2025tabtune}, while ensembling \tfms can damage calibration
\cite{tanna2026ensembling}. These benchmarks show that \tfms are competitive on average across many domains. They score
models with accuracy-type metrics and do not ask whether a gain translates into profit for a specific business
decision.

\paragraph{\tfms in business prediction.}
The closest analogue to our study is the credit-risk review of Baesens et al.\ \cite{baesens2026creditrisk}. Across
14 probability-of-default datasets TabPFN-3 ranks first, but its margin over the strongest GBDT is narrow and ``driven
by the latest TabPFN and TabICL generations rather than by the model class as a whole''. TabPFN-3 also underpredicts
the mean default rate. Tanna et al.\ found that, under heavy imbalance, building a class-balanced context adds 3 to 4 AUC
points over uniform sampling for subsampled contexts of 1k to 50k rows \cite{tanna2026resampling}; V4FinBench reports
a similar benefit of imbalance-aware context construction for bankruptcy prediction \cite{kostrzewa2026v4finbench}.
These studies show that the \tfm advantage depends on the model generation and that calibration and context
construction need their own checks; we test both for churn. For churn, the only \tfm benchmark we know of is
\cite{seyedzadeh2026churn}.

Table~\ref{tab:gap} shows why the three strands need to be combined. Manzoor et al.\ use tuned baselines and profit
metrics, but no \tfms. Baesens et al.\ compare \tfms with tuned GBDTs, but in credit risk and without a profit metric.
Seyedzadeh and Karimi study \tfms for churn, but without tuned baselines, a profit metric or significance tests. The
question an operator faces, whether a \tfm earns more campaign profit than the tuned GBDT it would otherwise use, needs
all three at once. Our contribution is therefore a test, not a new metric: we ask whether the reported \tfm advantage
for churn survives four conditions together. The conventional baselines are tuned; models are judged by a
campaign-oriented profit criterion; statistical uncertainty and multiple comparisons are accounted for; and
probability calibration and deployment terms are evaluated separately from ranking.

\begin{table}[t]
\caption{Closest prior studies. Tuned: baselines hyperparameter-tuned. Profit metric: a campaign profit measure used to
compare models. Calibration: probability calibration assessed with a dedicated measure (reliability diagram,
calibration error or calibration-in-the-large) or by recalibration; reporting log loss alone does not count, because it
mixes calibration with ranking. Tests: uncertainty or significance reported for differences between models.}
\label{tab:gap}
\centering\small
\resizebox{\textwidth}{!}{%
\begin{tabular}{@{}llccccc@{}}
\toprule
Study & Domain & \tfms & Tuned & Profit metric & Calibration & Tests \\
\midrule
Manzoor et al.\ \cite{manzoor2025eprofits} & churn & no & yes & e-Profits, EMP & no & yes \\
Baesens et al.\ \cite{baesens2026creditrisk} & credit risk & yes & yes & no & yes & yes \\
Seyedzadeh \& Karimi \cite{seyedzadeh2026churn} & churn & yes & no & no & log loss only & no \\
\midrule
This work & churn & yes & yes & EMP, e-Profits, policy & yes & yes \\
\bottomrule
\end{tabular}}
\end{table}

\section{Benchmark Design}

\subsection{Datasets}
Table~\ref{tab:data} lists the six datasets. We restricted the study to publicly available telecom datasets with a
binary churn label, so that every result can be recomputed by others; the selection is not exhaustive. Five datasets
form the main benchmark. Cell2Cell, Iranian and Orange come from
real operators: Cell2Cell from a US wireless carrier \cite{cell2cell}, Iranian from an Iranian operator through the UCI
repository \cite{iranian}, and Orange from the KDD Cup 2009 churn task \cite{guyon2009kddcup}, which we load from
OpenML (dataset 1112). IBM Telco \cite{telco} and Maven \cite{maven} are fictional teaching datasets; we keep them
because prior churn studies use them, including both \cite{manzoor2025eprofits} and \cite{seyedzadeh2026churn}, and
because a gap between real and fictional data would itself be informative. E-Commerce \cite{ecommerce}, the dataset on
which \cite{seyedzadeh2026churn} reported its largest gain, serves only as a replication check (Section~\ref{sec:rq5}) and
is excluded from all cross-dataset summaries. The three real-operator datasets carry most of the evidence. They differ
in size, number of features and churn rate (Table~\ref{tab:data}), so they cover different regimes, but three datasets are
too few to estimate how results depend on these properties, and none of them stands for telecom operators in general.

\begin{table}[t]
\caption{Datasets after preprocessing. Dup.: exact duplicate feature rows removed. e-Prof.: tenure and monthly revenue
available for e-Profits.}
\label{tab:data}
\centering\small
\resizebox{\textwidth}{!}{%
\begin{tabular}{@{}llrrrrcc@{}}
\toprule
Dataset & Source & Rows & Feat. & Churn \% & Dup. & e-Prof. & Role \\
\midrule
\input{tables/datasets.tex} 
\bottomrule
\end{tabular}}
\end{table}

All models share one preprocessing pipeline. Categorical columns are ordinal-encoded with missing values left as
missing, and columns that are more than 99\% missing or constant are dropped. For Maven we drop customers whose status
is ``Joined'', whose outcome is not yet observed, and the two columns that record the churn reason after the fact. We
remove exact duplicate feature rows, keeping the first, because an in-context learner can otherwise find a test
customer's twin in its own context. This removed \dupTelco{} rows (\dupPctTelco\%) from Telco, \dupIranian{}
(\dupPctIranian\%) from Iranian and \dupECommerce{} (\dupPctECommerce\%) from E-Commerce; the other datasets had none.
The removal changed churn rates by at most \maxChurnShift{} points, and Section~\ref{sec:sens} repeats the analysis with duplicates
kept.

\subsection{Models}
\textbf{Foundation models.} TabPFN-2.5, TabPFN-3.5 and TabPFN-3.5-Fast run through the \texttt{tabpfn} package (v9.2.0)
with default settings and the training-size guardrail disabled; TabICL v2 runs through \texttt{tabicl} (v2.2.0,
checkpoint of 2026-02-12). Each receives the full training split as its context, at the natural churn rate. None is
tuned, which is how these models are meant to be used.

\textbf{Tuned baselines.} LightGBM \cite{ke2017lightgbm} and XGBoost \cite{chen2016xgboost} are tuned with 40 trials
of Optuna's TPE sampler \cite{akiba2019optuna}, and CatBoost \cite{prokhorenkova2018catboost} with 25, each under a
20-minute cap. Trials are scored by validation AUC on an inner 80/20 split of the training data with early stopping.
We tune on AUC rather than EMP because AUC is a standard tuning objective and, like EMP, depends only on the ranking,
while EMP estimated on a small inner split, with few churners, is a noisier objective.
Because the primary metric is EMP, a GBDT tuned on EMP could be the stronger baseline and could narrow the EMP gaps.
We therefore repeat the tuning with each trial scored by EMP on the same inner split, with the same budgets, and
compare the \tfms against it as an exploratory check (Section~\ref{sec:sens}). The best
configuration is then refit on the whole training split with the selected number of trees. Search spaces are listed
in Appendix~\ref{app:repro}, and the configuration selected for every dataset, split, model and tuning objective is
released (Section~\ref{sec:availability}). We also report an explainable boosting machine (EBM) \cite{nori2019interpretml} and
logistic regression with one-hot encoding and median imputation, both at defaults.

\textbf{What the comparison is fair to.} The design matches how each model family would be used in practice rather
than giving every model the same compute. \tfms are used untuned, as intended; the three GBDTs, the strongest
conventional baselines for tabular data, are tuned; EBM and logistic regression are reference points and enter no
claim. A claim about ``tuned GBDTs'' in this paper therefore means these three models with these budgets, tuned on AUC.
The budgets differ (40 trials for LightGBM and XGBoost, 25 for CatBoost), and only LightGBM and CatBoost were rerun with
200 trials (Section~\ref{sec:sens}).

\subsection{Protocol and inference}\label{sec:inference}
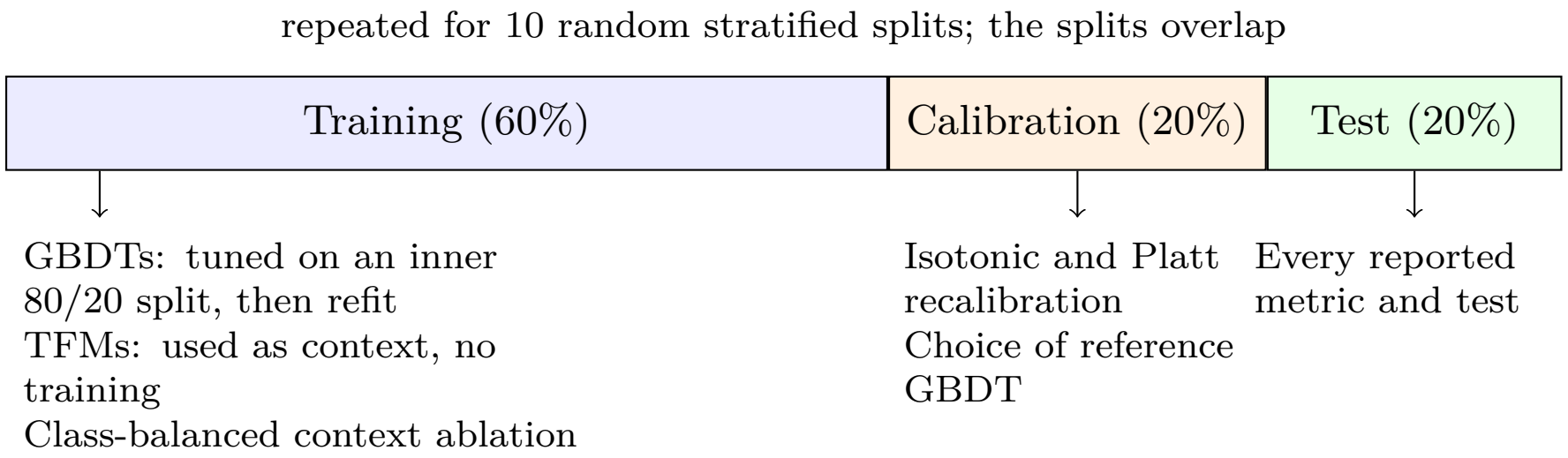
\begin{figure}[t]
\centering
\begin{tikzpicture}[font=\footnotesize, box/.style={draw, minimum height=7mm, align=center},
  use/.style={align=left, text width=4.2cm, font=\scriptsize}]
\node[box, fill=blue!8, minimum width=6.6cm] (tr) at (0,0) {Training (60\%)};
\node[box, fill=orange!12, minimum width=2.2cm, right=0mm of tr] (ca) {Calibration (20\%)};
\node[box, fill=green!10, minimum width=2.2cm, right=0mm of ca] (te) {Test (20\%)};
\node[use, below=4mm of tr.south west, anchor=north west] (u1) {GBDTs: tuned on an inner 80/20 split, then refit\\
\tfms: used as context, no training\\ Class-balanced context ablation};
\node[use, below=4mm of ca.south, anchor=north, text width=2.6cm] (u2) {Isotonic and Platt recalibration\\ Choice of
reference GBDT};
\node[use, below=4mm of te.south, anchor=north, text width=2.4cm] (u3) {Every reported metric and test};
\draw[->] (tr.south) ++(-2.6,0) -- ++(0,-0.35);
\draw[->] (ca.south) -- ++(0,-0.35);
\draw[->] (te.south) -- ++(0,-0.35);
\node[font=\scriptsize, anchor=south] at ($(tr.north west)!0.5!(te.north east)$) [yshift=1mm] {repeated for \nSeeds{} random stratified splits; the splits overlap};
\end{tikzpicture}
\caption{Use of each part of a split. The test part is touched only to report results. Because all \nSeeds{} splits
are drawn from the same customers, they overlap, and their results are correlated rather than independent.}
\label{fig:split}
\end{figure}

For each dataset and each of \nSeeds{} seeds we draw a stratified 60/20/20 split into training, calibration and test
sets (Figure~\ref{fig:split}). All models see identical splits, which the released audit verifies. Models are fitted
(or given context) on the training split. The calibration split has two uses: fitting post-hoc recalibration, and
choosing which tuned GBDT serves as the reference. Every reported evaluation metric is computed on the test split, which
plays no part in fitting, recalibration or the choice of reference. GBDTs ran on Kaggle CPU sessions and \tfms on a
single NVIDIA T4 (16\,GB).

\textbf{What is compared.} The estimand is the difference in a test-set metric between a \tfm and a baseline,
averaged over random splits. The \emph{primary family} is fixed in advance: EMP, each of the four \tfms against the best tuned GBDT on each of the five
main datasets, \nPrimary{} comparisons in all. The reference GBDT is chosen per dataset and metric as the tuned GBDT
(LightGBM, XGBoost or CatBoost) with the best mean of that metric on the calibration split over the \nSeeds{} splits; for
the primary family the metric is EMP. Choosing the reference on held-out data rather than on the test set prevents the
test results from deciding which baseline the \tfm is compared with. Two sensitivity analyses address the remaining
selection. Choosing the reference by its test mean instead is descriptive only: it
lets the test set influence the baseline, so it shows how much the choice matters but is not independent
confirmatory evidence. The second is an intersection--union test that requires a \tfm to beat each of the three tuned GBDTs separately, taking the largest of
the three $p$-values, again Holm-corrected over the \nPrimary{} comparisons.

\textbf{Why a corrected test.} Any two of the \nSeeds{} splits share most of their training customers and part of their
test customers, so their results move together. A standard $t$-test treats the \nSeeds{} results as independent,
understates the uncertainty and declares too many differences significant. The corrected resampled $t$-test of Nadeau
and Bengio \cite{nadeau2003inference} allows for the overlap by inflating the variance of the mean difference by
$1/J + n_\mathrm{test}/n_\mathrm{train}$ for $J$ splits. Even with this correction, the \nSeeds{} splits measure how
stable a comparison is within one dataset; they are not \nSeeds{} independent operators. Because the primary family
contains \nPrimary{} tests, we also apply Holm's correction \cite{holm1979simple}, which controls the chance of any false
positive across the family. We call a primary comparison significant when its Holm-adjusted $p$ is below 0.05, and report the
corrected test's 95\% confidence interval, which is unadjusted for multiplicity. A two-level bootstrap that resamples
seeds and then test customers is computed as a descriptive check; it does not model the overlap between splits, so it
is narrower than the corrected test, and no claim rests on it.
\textbf{Confirmatory and exploratory.} Only the \nPrimary{} primary comparisons support claims of significance.
Everything else, including PR-AUC, the intersection--union and test-selected sensitivity analyses, the context
ablation, the tuning-objective, tuning-budget and duplicate checks, the campaign-economics grid and the replication, is exploratory and
reported with unadjusted $p$.

\subsection{Metrics: three questions}\label{sec:metrics}
We keep three questions apart, because a model can answer one well and another badly.

\textbf{Does the model rank churners well?} ROC-AUC, PR-AUC and lift in the top decile.

\textbf{Does the ranking earn more campaign profit?} For a campaign that contacts the fraction of customers with the
highest scores, a contacted churner accepts the offer with probability $\gamma$ and is then worth
$\mathrm{CLV}-\delta-\phi$; a churner who declines costs the contact cost $\phi$; a contacted non-churner takes the
incentive and costs $\delta+\phi$. EMP is the expected profit per customer at the profit-maximising cutoff, averaged
over $\gamma\sim\mathrm{Beta}(6,14)$ (mean acceptance 0.3), with $\mathrm{CLV}=200$, $\delta=10$ and $\phi=1$, the
standard values of \cite{verbraken2013emp}. They are conventional defaults, not estimates for any operator in our
data, and Section~\ref{sec:scn} recomputes the comparison over \nScenarios{} combinations of CLV, incentive cost and
acceptance rate. We report EMP differences both per customer and relative to the baseline, as
$100\,(\overline{\mathrm{EMP}}_{\mathrm{TFM}}-\overline{\mathrm{EMP}}_{\mathrm{GBDT}})/\overline{\mathrm{EMP}}_{\mathrm{GBDT}}$,
where bars denote means over the compared splits; a small baseline EMP makes the relative figure large. Because EMP picks the best cutoff for each model, it depends only on the ranking: a high EMP does not mean the
probabilities can be used directly. We also report e-Profits \cite{manzoor2025eprofits}, which values each customer as
$\mathrm{CLV}_i = R_i M/(1-\min(r_i,0.995))$ from monthly revenue $R_i$, margin $M=0.3$ and one-month retention
$r_i = S(T_i+1)/S(T_i)$ from a Kaplan--Meier curve $S$ \cite{kaplan1958nonparametric} fitted on the training split. The
offer costs $0.1\,\mathrm{CLV}_i$ and the contact $\max(5, 0.3\cdot\text{offer})$. We report it when contacting the top
20\% and, as in \cite{manzoor2025eprofits}, when contacting customers with predicted probability at least 0.5; the
latter mixes ranking with calibration. EMP is our primary profit measure because it needs no customer-level revenue,
which Iranian and Orange do not provide, and because it averages over uncertainty about offer acceptance. e-Profits is
available for three datasets only and is supplementary. Both are model-based estimates of campaign profit under
assumed economics; neither is profit measured in a real campaign.

\textbf{Can the probabilities drive a threshold policy?} An operator who acts on probabilities contacts customer $i$
when its expected profit under the EMP cost model is positive at the mean acceptance rate $\bar\gamma=0.3$. We report
the policy's regret: the profit lost relative to the best cutoff on the same ranking, as a share of that best profit,
as the median over a $3\times3$ grid of $\mathrm{CLV}\in\{100,200,500\}$ and $\delta\in\{5,10,20\}$. Calibration is
measured three ways, which answer different questions. Calibration-in-the-large (CITL) asks whether the overall
predicted churn rate is right: it is the observed churn rate minus the mean predicted probability, in percentage
points, positive when churn is underpredicted. Expected calibration error (ECE, 15 equal-width bins) asks whether
individual probability levels are right: it averages the gap between predicted and observed churn within bins. The
Brier score, the mean squared error of the probabilities, mixes calibration with ranking quality, so a low Brier score
does not by itself show good calibration.
Each model is evaluated raw and after isotonic regression \cite{zadrozny2002transforming} or Platt scaling
\cite{platt1999probabilistic}, both fitted on the calibration split.

\section{Results}

\subsection{Do foundation models earn more profit than tuned GBDTs?}\label{sec:rq1}

\begin{figure}[t]
\centering
\includegraphics[width=0.78\textwidth]{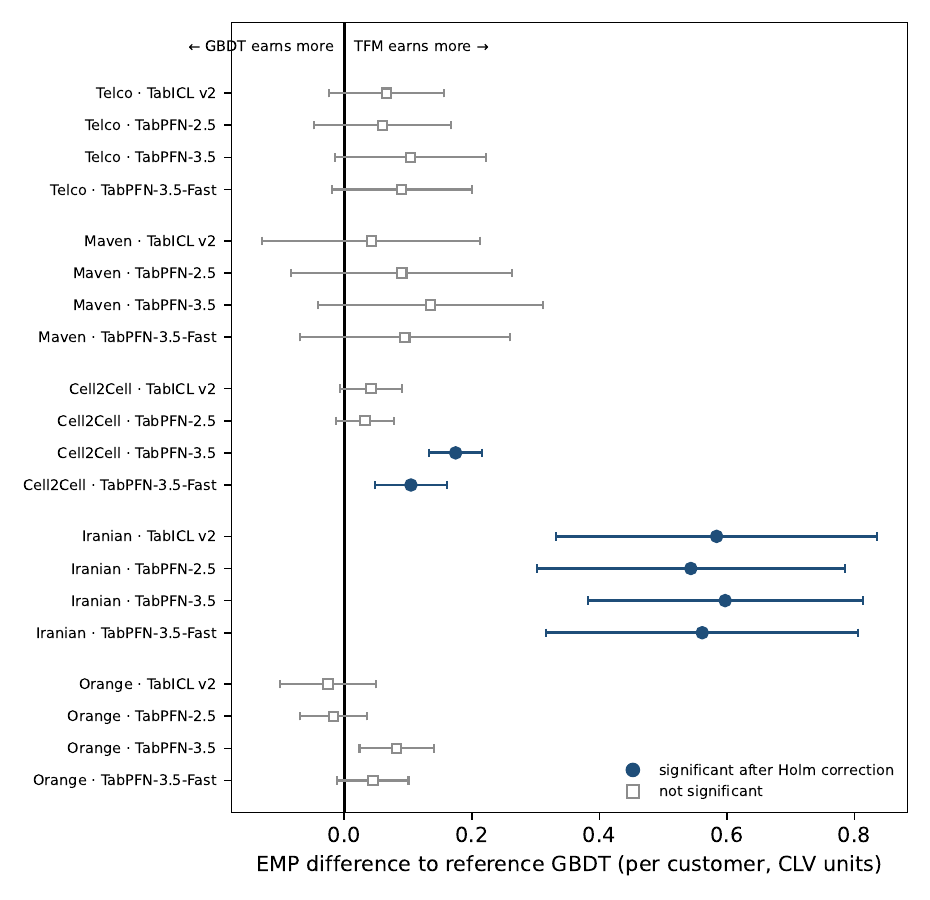}
\caption{EMP difference between each \tfm and the reference GBDT, per customer, over \nSeeds{} splits. Points to the
right of the zero line mean the \tfm selects a more profitable contact list. Bars are 95\% intervals from the corrected
$t$-test, not adjusted for the \nPrimary{} comparisons. Dark filled circles: significant after Holm correction; grey open
squares: not significant. An interval can exclude zero and still not be significant after correction (Orange,
TabPFN-3.5).}
\label{fig:forest}
\end{figure}

TabPFN-3.5 earns significantly more EMP than the reference GBDT on two of five datasets, Cell2Cell and Iranian
(Table~\ref{tab:primary}, Figure~\ref{fig:forest}). On Cell2Cell it selects customers worth \empAbsPFNThreeCellTwoCellAbs{}
more per customer in the base, \empPFNThreeCellTwoCellAbs\% of the reference GBDT's EMP (Holm $p$
\holmPFNThreeCellTwoCell); on Iranian the gain is \empAbsPFNThreeIranianAbs{} per customer, or \empPFNThreeIranianAbs\%
(Holm $p$ = \holmPFNThreeIranian). Both gains also hold against each of the three tuned GBDTs separately. In total,
\nHolmSig{} of the \nPrimary{} primary comparisons are significant after correction, and none favours a GBDT.

\textbf{Against the best GBDT and against all three.} The primary family asks whether a \tfm beats the GBDT that looked
best on held-out data. The stricter intersection--union test asks whether it beats every tuned GBDT, and only \nIUSig{}
comparisons pass it: \iuSigCases. On Iranian all four \tfms are significantly ahead of the reference GBDT, but for three
of them the result depends on which GBDT is the baseline. We therefore claim superiority over tuned GBDTs as a group
only for these \nIUSig{} comparisons. Choosing the reference on test means, a descriptive check that is not confirmatory
because it uses the test set, also gives \nHolmSigTest{} significant comparisons.

\textbf{Three levels of evidence.} A result can mean three different things, and we keep them apart. A \tfm can be
ahead in the observed splits; the difference can be significant under the stated procedure; and the difference can be
large enough to matter to an operator. The first two are statistical statements about these datasets. The third
depends on the operator's economics. Under the assumed economics, the Cell2Cell gain amounts to about
\perMillionPFNThreeCellTwoCell{} currency units per campaign for an operator with one million customers, against a
CLV of 200 per retained customer. Whether that justifies a change of model depends on the operator's actual values,
the cost of switching and the licence terms discussed in Section~\ref{sec:rq6}.

\textbf{Orange illustrates the effect of multiple-comparison correction most clearly.} TabPFN-3.5 is ahead of the
reference GBDT in all \nSeeds{} splits, by \empAbsPFNThreeOrangeAbs{} per customer. That absolute gain is smaller than on
Cell2Cell. The relative figure, \empPFNThreeOrangeAbs\%, is large because the reference EMP is small
(\refEMPPFNThreeOrange{} per customer): only \rateOrange\% of Orange customers churn. Relative to its size, the gain
also varies more across splits than on Cell2Cell. Its corrected $p$ is \nbpPFNThreeOrange{} before adjustment and
\holmPFNThreeOrange{} after, so under the primary procedure the study does not establish an advantage on Orange. We read
it as a consistent direction that warrants further study, not as a confirmed benefit. The relative figure in particular
should not be carried to other operators, because the \nSeeds{} splits resample one operator's customers.

On the two fictional datasets no difference survives correction. TabPFN-3.5 is ahead by \empPFNThreeTelcoAbs\% on Telco
and \empPFNThreeMavenAbs\% on Maven, with Holm-adjusted $p$ of 1. TabICL v2 beats the reference GBDT only on Iranian, and
on Orange its mean EMP is below the GBDT's (\empICLOrange\%, not significant). PR-AUC differences, treated as exploratory,
point the same way (Appendix, Table~\ref{tab:boot}).

\textbf{Why results differ by dataset.} Five datasets are too few to test which property drives the differences, so
what follows are hypotheses consistent with the data rather than findings. The largest and most uniform gains occur on
Iranian, the smallest dataset (Table~\ref{tab:data}), where all four \tfms lead the reference GBDT by similar margins. This
matches the TabArena observation that \tfms have their largest advantage on small datasets \cite{erickson2025tabarena},
where a pretrained prior can stand in for data a GBDT must learn from. Cell2Cell and Orange both have about 50k customers. On
Cell2Cell only the two TabPFN-3.5 models are significantly ahead, and on Orange TabPFN-3.5 is the only model ahead in
every split, which matches the credit-risk finding that the newest
generation, rather than the class, drives most of the gain \cite{baesens2026creditrisk}. TabPFN-2.5 and TabPFN-3.5 share
a lineage, yet only the latter separates from the GBDTs on Cell2Cell. Orange also has the most features and the lowest
churn rate of the five datasets; TabICL v2's weaker result there may relate to that imbalance (Section~\ref{sec:rq3}), but
we did not test this. We cannot tell whether the null results on Telco and Maven reflect how those fictional datasets
were generated or their moderate size.

\begin{table}[t]
\caption{Primary family: EMP of each \tfm minus the reference GBDT, per
customer (CLV units) and as a percentage of the GBDT's mean EMP over the same splits (Section~\ref{sec:metrics}). CI: 95\% interval of the corrected resampled $t$-test, unadjusted for
multiplicity. Seeds: splits in which the \tfm is ahead. $p$: corrected test; Holm: adjusted over the \nPrimary{}
comparisons. All: Holm-adjusted intersection--union $p$ against each of the three tuned GBDTs. Bold: below 0.05.
Significance is decided only by the Holm-adjusted columns; an unadjusted CI that excludes zero is not a claim.
Reference: the tuned GBDT with the best mean calibration-split EMP, and its mean test EMP per customer.}
\label{tab:primary}
\centering\scriptsize\setlength{\tabcolsep}{3.2pt}
\begin{tabular}{@{}lllrrccccc@{}}
\toprule
Dataset & Reference & \tfm & $\Delta$EMP & \% & 95\% CI (corrected) & Seeds & $p$ & Holm & All \\
\midrule
\input{tables/primary.tex} 
\bottomrule
\end{tabular}
\end{table}

\subsection{Do profit metrics change which model wins?}\label{sec:rq2}

They change it when models are tied and leave it alone otherwise (Table~\ref{tab:rank}). On Cell2Cell and Orange, where
TabPFN-3.5 leads clearly, the EMP winner differs from the AUC winner in only \topEMPCellTwoCell{} and \topEMPOrange{} of \nSeeds{} splits. On Telco and Maven,
where no difference is significant, the EMP winner differs from the AUC winner in \topEMPTelco{} and \topEMPMaven{} of
\nSeeds{} splits. Iranian sits in between (\topEMPIranian{} of \nSeeds): the four \tfms are tied with each other there,
though all lead the GBDTs.

e-Profits at the 0.5 threshold disagrees with AUC the most ($\tau=\tauEPthrMaven$ on Maven). This follows from its
construction rather than from a better ranking: at a fixed threshold the metric also rewards a model whose
probabilities happen to fall on the right side of 0.5. The reranking reported for e-Profits on Telco and Maven
\cite{manzoor2025eprofits} is therefore reproduced here, among models whose rankings are statistically
indistinguishable.

\begin{table}[t]
\caption{Agreement between the AUC ranking of all nine models and profit-based rankings: mean Kendall $\tau$ over
splits, and splits in which the top model changes. e-Profits needs tenure and revenue.}
\label{tab:rank}
\centering\small
\begin{tabular}{@{}lcccccc@{}}
\toprule
& \multicolumn{2}{c}{EMP} & \multicolumn{2}{c}{e-Profits, top 20\%} & \multicolumn{2}{c}{e-Profits, $p\ge0.5$} \\
\cmidrule(lr){2-3}\cmidrule(lr){4-5}\cmidrule(l){6-7}
Dataset & $\tau$ & top-1 changed & $\tau$ & top-1 changed & $\tau$ & top-1 changed \\
\midrule
\input{tables/rank_agreement.tex} 
\bottomrule
\end{tabular}
\end{table}

\subsection{Can the probabilities drive a threshold policy?}\label{sec:rq3}

Ranking quality and probability reliability are different properties. EMP uses only the order of customers, so a
model can rank churners well and still give probabilities that are too high or too low. That matters as soon as an
operator acts on the probabilities themselves, for example by contacting every customer whose expected profit is
positive at a fixed cutoff. If churn is underpredicted, fewer customers clear the cutoff and churners worth contacting
are missed; if it is overpredicted, the campaign contacts customers who were never at risk.

For most models the probabilities are reliable enough. Outside Orange, the fixed-threshold policy loses at most
\maxRawRegretNonOrange\% of attainable profit without recalibration, and isotonic or Platt scaling changes that loss by at
most \maxRegretShiftNonOrange{} points (Table~\ref{tab:cal}).

\begin{figure}[t]
\centering
\includegraphics[width=\textwidth]{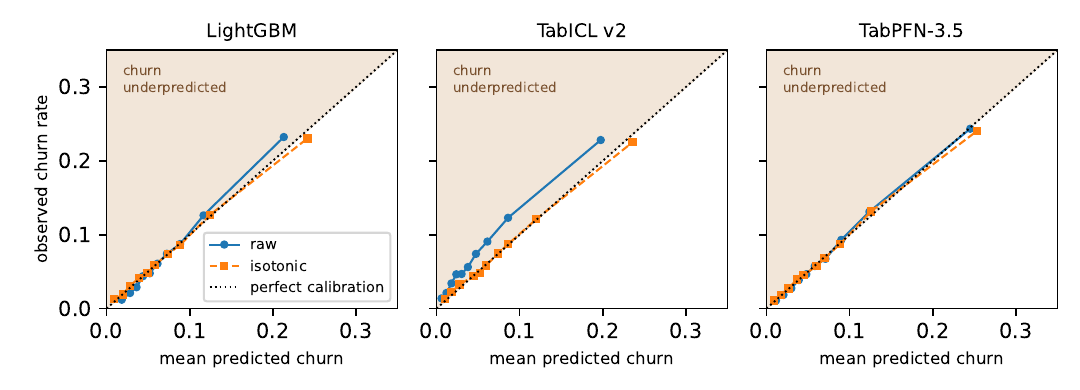}
\caption{Reliability on Orange, test customers pooled over \nSeeds{} splits, ten quantile bins. A perfectly calibrated
model lies on the dotted diagonal: among customers given a 20\% churn probability, 20\% churn. Points in the shaded
region above it mean churn is underpredicted. Raw TabICL v2 lies above the diagonal; isotonic recalibration on the
calibration split removes the gap. Pooling the \nSeeds{} test splits shows the average pattern; the pooled customers
overlap across splits and are not an independent validation set. Table~\ref{tab:cal} reports calibration measures computed per split and averaged.}
\label{fig:rel}
\end{figure}

Orange is the exception, and one model accounts for it. Raw TabICL v2 predicts \citlICLOrangeAbs{} percentage points too
little churn on a base rate of \rateOrange\% (Figure~\ref{fig:rel}), and the policy built on those probabilities loses
\regICLOrangeRaw\% of attainable profit. Isotonic regression cuts the loss to \regICLOrangeIso\% and Platt scaling to
\regICLOrangePlatt\%. TabICL v2 also underpredicts on Cell2Cell (by \citlICLCellTwoCellAbs{} points), as does TabPFN-2.5
(\citlPFNTwoCellTwoCellAbs). Among the evaluated \tfms, TabPFN-3.5 and TabPFN-3.5-Fast show the most consistent calibration-in-the-large across the
five datasets, within $\pm$\maxAbsCitlPFNThree{} and $\pm$\maxAbsCitlPFNFast{} points, with comparatively low regret under the
tested threshold policies. This does not establish that their raw probabilities are reliable for every operational
threshold or for other operators' data. The tuned GBDTs are close to zero as well, consistent with their fitting to the observed
base rate. Among all nine models, TabPFN-3.5 has the lowest raw Brier score on \nBrierBestPFNThree{} of five datasets and the
lowest raw ECE on \nECEBestPFNThree. We do not read these as evidence of the best calibration: the Brier score also
rewards TabPFN-3.5's better ranking, and the ECE differences between the better-calibrated models are small. The
practically relevant result is the CITL pattern above, together with the policy regret it produces.

\begin{table}[t]
\caption{Calibration of raw and recalibrated probabilities. CITL: observed minus mean predicted churn rate (points;
positive = underprediction). Brier: raw Brier score. Regret: share of best-cutoff profit lost by the fixed-threshold
policy (median over the cost grid), raw / isotonic / Platt. The first row per dataset is the reference GBDT (best tuned
GBDT on calibration-split EMP); the other two tuned GBDTs, EBM and logistic regression are omitted for space, and all
models' calibration metrics are in the released \texttt{metrics\_summary.csv}.}
\label{tab:cal}
\centering\scriptsize
\begin{tabular}{@{}llrrrrrrr@{}}
\toprule
& & & & \multicolumn{2}{c}{ECE} & \multicolumn{3}{c}{Policy regret (\%)} \\
\cmidrule(lr){5-6}\cmidrule(l){7-9}
Dataset & Model & CITL & Brier & raw & iso. & raw & iso. & Platt \\
\midrule
\input{tables/calibration.tex} 
\bottomrule
\end{tabular}
\end{table}

\subsection{Does a balanced context help?}\label{sec:rq4}

Under heavy imbalance, a balanced context has been reported to add 3 to 4 AUC points for \tfms
\cite{tanna2026resampling}. If that held here, our full-context runs would understate the \tfms on Orange. For
TabPFN-3.5 and TabICL v2 on Orange and Cell2Cell we kept every churner in the training split and sampled one or two
non-churners per churner, on the first five splits.

Balanced context construction reduced the observed advantage (Table~\ref{tab:ctx}). TabPFN-3.5's EMP gain on Orange fell from \ctxOrangePFNFullAbs\% with the full
context to \ctxOrangePFNHalfAbs\% at 1:2 and \ctxOrangePFNBalAbs\% at 1:1 (unadjusted $p$ \ctxOrangePFNpHalf{} and
\ctxOrangePFNpBal). TabICL v2 gained nothing on either dataset. Balancing also inflated raw predicted churn, by
\ctxCitlOrangeAbs{} points on Orange at 1:1, which isotonic recalibration removes. The difference from
\cite{tanna2026resampling} is the regime: there, contexts were subsampled to at most 50k rows from larger data, so
balancing chose which customers to keep. Here the whole training split fits, and undersampling only discards
non-churners.

\begin{table}[t]
\caption{Context construction (exploratory). Full: the whole training split at its natural churn rate (\nSeeds{}
splits). 1:2 and 1:1: all churners plus two or one sampled non-churners per churner (five splits). $\Delta$EMP relative
to the best tuned GBDT, splits ahead, and unadjusted corrected $p$. CITL in points, raw.}
\label{tab:ctx}
\centering\scriptsize
\resizebox{\textwidth}{!}{%
\begin{tabular}{@{}lllrrrrccr@{}}
\toprule
Dataset & Model & Context & AUC & PR-AUC & EMP & $\Delta$EMP & Seeds & $p$ & CITL \\
\midrule
\input{tables/context.tex} 
\bottomrule
\end{tabular}}
\end{table}

\subsection{Robustness to tuning objective, tuning budget and duplicates}\label{sec:sens}

All three checks below are exploratory, and their $p$-values are unadjusted. The tuning-objective check covers every
split of the five primary datasets; the tuning-budget and duplicate checks cover five splits of three datasets, not the
full primary design. They can show that a conclusion did not change where they were run; they cannot show that it would
not change elsewhere.

\textbf{Tuning objective (all \nSeeds{} splits of the five datasets).} The primary GBDTs are tuned on AUC, while the
comparison is on EMP. We retuned LightGBM, XGBoost and CatBoost with each trial scored by EMP on the same inner split
and with the same budgets, and chose the reference on the calibration split among all six tuned GBDTs, AUC- and
EMP-tuned. The EMP-tuned model became the reference on \nEmpRefDatasets{} datasets (\empRefDatasets). Across the
\nPrimary{} comparisons, no \tfm's mean gain moved by more than \maxEmpTuneShift{} profit units per customer, and
\nEmpTuneSignFlip{} changed sign (Table~\ref{tab:emptune}). On Iranian and Orange the EMP-tuned GBDTs were weaker on
the calibration split, so the reference is unchanged; against the best EMP-tuned GBDT alone, TabPFN-3.5's gain is
larger (\empTunedPFNThreeIranianAbs\% on Iranian and \empTunedPFNThreeOrangeAbs\% on Orange). On Maven, TabPFN-3.5's gain over the
EMP-tuned CatBoost is slightly smaller than over the primary reference, but its split-level differences vary less, so
the unadjusted $p$ falls to \anyTunedPFNThreeMavenP; this is an exploratory, unadjusted value and not a significance
claim. As with the larger
budget, this shows that tuning on the primary metric did not produce a stronger baseline within this search space and
budget, not that no profit-aware tuning could.

\textbf{Tuning budget.} Our baselines use 25 to 40 trials, fewer than benchmark practice \cite{erickson2025tabarena}.
We reran LightGBM and CatBoost with 200 trials each on Cell2Cell, Iranian and Orange for the first five splits only,
and compare on those five splits throughout this check: both references are chosen on the calibration split over them,
and both columns of Table~\ref{tab:sens}a are computed on them. The larger budget barely moved the best GBDT's mean test EMP: by $\strongDeltaCellTwoCell\%$ on Cell2Cell,
$\strongDeltaIranian\%$ on Iranian and $\strongDeltaOrange\%$ on Orange. On Cell2Cell the
200-trial LightGBM becomes the reference, against which TabPFN-3.5 retains a mean EMP gain of
\strongPFNThreeCellTwoCellAbs\% and is ahead in all five splits. On Iranian and Orange no 200-trial model beats the existing reference on the calibration split, so the
reference and the gains are unchanged; this part of the check shows that the extra budget did not produce a better
baseline, not that a stronger baseline was beaten. The five-split gains differ from the ten-split primary figures
(for example \strongPFNThreeOrangeAbs\% rather than \empPFNThreeOrangeAbs\% on Orange) because they cover half the
splits, and their $p$-values are unadjusted.

\textbf{Duplicates (five splits of Telco, Iranian and E-Commerce).} Removing duplicates could discard legitimate
customers. A duplicate group is a set of rows with identical
features; removal keeps one row per group, discarding the other copies. In E-Commerce every group's copies share the
same churn label. In Iranian, \conflictGroupsIranian{} groups (\conflictRowsIranian{} rows in total) and in Telco
\conflictGroupsTelco{} groups (\conflictRowsTelco{} rows) contain copies with different labels, so keeping the first
copy assigns them one label; removal changed churn rates by at most
\maxChurnShift{} points. We repeated the analysis keeping every row but assigning all copies of a feature vector to the same side of
each split, so that no twin crosses from context to test. The gains moved by at most \maxDupShift{} points on all three datasets
(Table~\ref{tab:sens}b).

\begin{table}[t]
\caption{Tuning objective (exploratory, \nSeeds{} splits). Mean $\Delta$EMP of each \tfm, in profit units per customer
and relative to the reference's mean EMP, against the primary reference (best GBDT tuned on AUC, named in
Table~\ref{tab:primary}) and against the best of the AUC- and EMP-tuned GBDTs, both chosen on the calibration split.
(EMP): the reference was tuned on EMP. Seeds: splits in which the \tfm is ahead of the right-hand reference. $p$:
unadjusted corrected $t$-test for the right-hand comparison.}
\label{tab:emptune}
\centering\scriptsize
\begin{tabular}{@{}llrrlrrcc@{}}
\toprule
 & & \multicolumn{2}{c}{AUC-tuned reference} & \multicolumn{5}{c}{Best of AUC- and EMP-tuned} \\
\cmidrule(lr){3-4}\cmidrule(l){5-9}
Dataset & \tfm & $\Delta$EMP & \% & Reference & $\Delta$EMP & \% & Seeds & $p$ \\
\midrule
\input{tables/emptuned.tex} 
\bottomrule
\end{tabular}
\end{table}

\begin{table}[t]
\caption{Sensitivity (exploratory). (a) On the first five splits, where the 200-trial runs exist: $\Delta$EMP of each
\tfm against the best GBDT tuned with 25--40 trials, and against the best of those and 200-trial LightGBM and CatBoost.
Both references are chosen on the calibration split over the same five splits. (b) $\Delta$EMP with duplicates removed (primary
protocol) and with duplicates kept but split by duplicate group (five splits). Removed: copies discarded by de-duplication; Confl.:
duplicate groups whose copies disagree on churn, with the rows they contain. $p$: unadjusted corrected $t$-test for
the right-hand comparison.}
\label{tab:sens}
\centering\scriptsize
\begin{tabular}{@{}lllrlrcc@{}}
\toprule
\multicolumn{8}{@{}l}{\emph{(a) Tuning budget}} \\
Dataset & \tfm & Ref.\ (25--40 trials) & $\Delta$EMP & Ref.\ (incl.\ 200 trials) & $\Delta$EMP & Seeds & $p$ \\
\midrule
\input{tables/strong.tex} 
\bottomrule
\end{tabular}

\medskip
{\setlength{\tabcolsep}{4.5pt}\begin{tabular}{@{}lrrclrrc@{}}
\toprule
\multicolumn{8}{@{}l}{\emph{(b) Duplicates}} \\
Dataset & Removed & Confl.\ groups (rows) & Churn \% (before/after) & \tfm & Dedup. & Grouped & $p$ \\
\midrule
\input{tables/dupgroup.tex} 
\bottomrule
\end{tabular}}
\end{table}

\subsection{Do the conclusions depend on the assumed campaign economics?}\label{sec:scn}

The EMP comparison rests on assumed values of customer lifetime value, incentive cost and acceptance rate. We
recomputed it under \nScenarios{} combinations: CLV of 100, 200 or 500; incentive cost $\delta$ of 5, 10 or 20; and a
mean acceptance rate of 0.1, 0.3 or 0.5 (Beta(2,18), Beta(6,14) and Beta(10,10)), with the contact cost fixed at 1. For
each combination the reference GBDT is re-chosen on calibration-split EMP under that combination, so the comparison is
redone rather than rescaled. This analysis is exploratory, and its $p$-values are unadjusted.

Within the tested grid, the direction of TabPFN-3.5's advantage on the three real-operator datasets does not depend on
the assumptions: it is ahead of the reference GBDT in \scnPosCellTwoCell{}, \scnPosIranian{} and \scnPosOrange{} of the \nScenarios{}
combinations on Cell2Cell, Iranian and Orange (Table~\ref{tab:scn}). The size of the advantage does depend on them. It
shrinks towards zero when customers are valuable and the incentive cheap or likely to be accepted, because the
profit-maximising campaign then contacts almost everyone and the order of the ranking matters little. The largest
relative figures should not be read as large financial gains. They occur where the reference GBDT earns almost nothing:
the \scnRelMaxOrangeAbs\% maximum on Orange is an absolute gain of \scnAbsMaxOrange{} per customer on a reference EMP of
\scnRefMaxOrange, about the same absolute gain as in Orange's lowest-percentage combination (Table~\ref{tab:scn}), and the
\scnRelMaxCellTwoCellAbs\% maximum on Cell2Cell is \scnAbsMaxCellTwoCell{} per customer. The grid covers \nScenarios{}
chosen combinations, not all plausible economics; an operator should recompute EMP with its own values.

On the fictional datasets the assumptions matter more. The model with the highest EMP on Telco is the same as under the
default economics in only \scnTopSameTelco{} of \nScenarios{} combinations, consistent with the near-ties in
Section~\ref{sec:rq2}. Across all 20 \tfm--dataset pairs, \nScnSignFlip{} change the sign of their mean difference
somewhere on the grid; all of them are pairs whose primary comparison is not significant.

The grid does not change the confirmatory result for Orange. TabPFN-3.5's difference there has unadjusted $p<0.05$ in
\scnSigOrange{} of \nScenarios{} combinations, but so does the default comparison ($p$ = \nbpPFNThreeOrange), which does
not survive Holm correction. The grid shows that the direction of the Orange gain is stable across economics, not that
it is significant.

\begin{table}[t]
\caption{Campaign-economics sensitivity (exploratory): \nScenarios{} combinations of CLV, incentive cost and mean
acceptance rate. Top model: highest mean EMP among all nine models under the default economics, and the number of
combinations in which it stays on top. For TabPFN-3.5 against the reference GBDT re-chosen per combination: the
combinations with the lowest and the highest relative gain, each with the relative gain, the absolute gain per
customer and the reference GBDT's EMP; then the combinations in which TabPFN-3.5 is ahead and those with unadjusted
corrected $p<0.05$. A large percentage on a small reference EMP can be a small absolute gain.}
\label{tab:scn}
\centering\scriptsize\setlength{\tabcolsep}{3.2pt}
\begin{tabular}{@{}llcrrrrrrcc@{}}
\toprule
& & & \multicolumn{3}{c}{Lowest gain} & \multicolumn{3}{c}{Highest gain} & & \\
\cmidrule(lr){4-6}\cmidrule(lr){7-9}
Dataset & Top model & Stays top & \% & $\Delta$EMP & Ref.\ EMP & \% & $\Delta$EMP & Ref.\ EMP & Ahead & $p<0.05$ \\
\midrule
\input{tables/scenarios.tex} 
\bottomrule
\end{tabular}
\end{table}

\subsection{Does the previously reported advantage persist after baseline tuning?}\label{sec:rq5}

We reran our protocol on E-Commerce, where \cite{seyedzadeh2026churn} reported TabICL v2 9.23 PR-AUC points ahead of
XGBoost, adding a library-default XGBoost to stand in for their baseline. This is a replication under a different
protocol, not a reproduction (Table~\ref{tab:protocols}). TabICL v2 leads default XGBoost by \replICLDefaultAbs{}
points and the best tuned GBDT, \replTuned, by \replICLTunedAbs{} points, ahead in all \replICLSeeds{} splits (unadjusted $p$
\replICLTunedP), with an EMP gain of \replICLEMPAbs\% (Table~\ref{tab:repl}). Tuning raised the GBDT's PR-AUC from
\replDefaultPR{} to \replTunedPR{} and so removed about a third of the gap. TabPFN-3.5 leads by more
(\replPFNThreeTunedAbs{} points over tuned \replTuned; EMP gain \replPFNThreeEMPAbs\%). On this dataset, then, \tfms
win clearly. The origin of E-Commerce is not documented, so it cannot tell us whether the effect is specific to real
or synthetic data.

\begin{table}[t]
\caption{Protocols of the earlier benchmark \cite{seyedzadeh2026churn} and of this replication on E-Commerce.}
\label{tab:protocols}
\centering\scriptsize
\begin{tabular}{@{}p{0.16\textwidth}p{0.36\textwidth}p{0.42\textwidth}@{}}
\toprule
& Seyedzadeh \& Karimi \cite{seyedzadeh2026churn} & This work \\
\midrule
Splits & 5-fold stratified CV & 60/20/20 holdout, 5 seeds \\
Duplicates & not stated & removed (\dupECommerce{} rows) \\
GBDT baseline & XGBoost, library defaults & tuned LightGBM, XGBoost, CatBoost; default XGBoost \\
\tfms & TabICL v1 and v2, TabPFN v2, TabDPT, Mitra, LimiX, ConTextTab & TabICL v2, TabPFN-2.5, 3.5, 3.5-Fast \\
Metrics & F1, ROC-AUC, PR-AUC, log loss & PR-AUC, EMP, with corrected $t$-test \\
\bottomrule
\end{tabular}
\end{table}

\begin{table}[t]
\caption{Replication on E-Commerce (exploratory, five splits). PR-AUC mean (sd); $\Delta$ in PR-AUC points versus
default XGBoost and versus the best tuned GBDT, and EMP relative to the best tuned GBDT. $\dagger$: two-level bootstrap
interval excludes zero.}
\label{tab:repl}
\centering\scriptsize
\begin{tabular}{@{}lccccc@{}}
\toprule
Model & PR-AUC & EMP & $\Delta$PR vs default & $\Delta$PR vs tuned & $\Delta$EMP \\
\midrule
\input{tables/replication.tex} 
\bottomrule
\end{tabular}
\end{table}

\subsection{Scoring cost and licensing}\label{sec:rq6}

An in-context learner re-reads its context for every scoring call unless the context is cached, so its cost depends on
how scoring is run. We timed scoring of 1k, 10k and 50k customers resampled from the test split, after a warm-up call,
on one T4 (Table~\ref{tab:time}). With the context cached, TabPFN-3.5 scores 10k Orange customers in
\timePFNCacheOrange{}\,s against a \ctxOrange-row context, and TabICL v2 with its key-value cache in
\timeICLCacheOrange{}\,s, against \timeLGBMOrange{}\,s for a 1,000-tree LightGBM on CPU. A monthly scoring run of a
million customers would therefore take about \hoursPFNCacheMillion{} T4-hours with cached TabPFN-3.5. This conclusion is
conditional on batch scoring, a cached context and this hardware; real-time serving with tight latency budgets is a
different question we did not measure. Without caching, small batches are very expensive
(\timePFNUncachedSmallOrange\,s per 10k for TabPFN-3.5 at 1k rows on Orange), and TabPFN-2.5 was impractical at
30k-row contexts.

The binding constraint is the licence. The TabPFN-3.5 weights (including the -Fast checkpoint, distributed in the same repository) and the TabPFN-2.5
weights are released under non-commercial
licences that forbid using the model, its derivatives ``or any data produced by'' the model ``for any commercial or
production purposes'' \cite{tabpfn35license,tabpfn25license}; the licences define outputs to include predictions,
scores and probabilities. The TabPFN-3.5 report states the practical consequence: outputs may not be used ``as inputs
to internal commercial decision-making'' without a commercial licence \cite[Section~4]{priorlabs2026tabpfn35}. The
TabICL v2 classifier checkpoint we used (\texttt{tabicl-classifier-v2-20260212}) is distributed under BSD-3
\cite{tabiclweights}. These terms were read on 2026-10-28 and may change, as may the checkpoints themselves; the
conclusions here apply to the versions listed in Appendix~\ref{app:repro}. An
operator therefore faces a trade-off that our results make concrete: the model with the most consistent gains needs a
commercial agreement, and the openly licensed one beats the best tuned GBDT on only one of five datasets, and not all
three tuned GBDTs on any.

\begin{table}[t]
\caption{Seconds to score 10k customers on one T4 (LightGBM on CPU), best of 10k and 50k batches. $\dagger$: only the
1k batch was run because larger ones exceeded our time budget; per-customer time falls with batch size, so this is an
upper bound. Context: training split (\ctxTelco{} / \ctxCellTwoCell{} / \ctxOrange{} rows).}
\label{tab:time}
\centering\small
\begin{tabular}{@{}lrrr@{}}
\toprule
Model & Telco & Cell2Cell & Orange \\
\midrule
\input{tables/timing.tex} 
\bottomrule
\end{tabular}
\end{table}

\section{Discussion}

The results support a narrower statement than ``foundation models outperform trees for churn''. With tuned baselines,
a profit criterion and a correction for multiple comparisons, one model generation, TabPFN-3.5, earns more expected
campaign profit than each of three tuned GBDTs on two of five datasets and is never significantly behind. The other
\tfms are rarely distinguishable from a tuned GBDT. Three factors account for most of the variation we observe. The
model version matters more than the model class. Tuning the baseline removes part of any measured lead, about a third
on E-Commerce. And the two datasets most used in churn research, Telco and Maven, are those on which no model can be
told apart, so studies built on them can neither confirm nor refute a \tfm advantage.

Table~\ref{tab:evidence} sorts the findings by the strength of the evidence behind them. Only the first two rows rest
on the confirmatory procedure with the stricter test against every tuned GBDT; everything below the line is either not
significant or exploratory.

\begin{table}[t]
\caption{Evidence status of the main findings. Holm: Holm-adjusted $p$ over the \nPrimary{} primary comparisons; IU:
Holm-adjusted intersection--union $p$ against each of the three tuned GBDTs.}
\label{tab:evidence}
\centering\scriptsize
\begin{tabular}{@{}p{0.25\textwidth}p{0.38\textwidth}p{0.31\textwidth}@{}}
\toprule
Finding & Evidence status & Interpretation \\
\midrule
TabPFN-3.5 on Cell2Cell and Iranian & Primary; Holm $p$ \holmPFNThreeCellTwoCell{} and \holmPFNThreeIranian; IU $p$
\iuHolmPFNThreeCellTwoCell{} and \iuHolmPFNThreeIranian & Strongest supported result \\
TabPFN-3.5-Fast on Cell2Cell & Primary; Holm $p$ \holmPFNFastCellTwoCell; IU $p$ \iuHolmPFNFastCellTwoCell & Supported \\
\midrule
Other \tfms on Iranian & Primary; Holm-significant against the reference GBDT, not significant against all three & Ahead
of the selected GBDT, not of every tuned GBDT \\
TabPFN-3.5 on Orange & Primary; ahead in \seedsPosPFNThreeOrange/\nSeeds{} splits; Holm $p$ \holmPFNThreeOrange & Consistent
direction, inconclusive \\
All \tfms on Telco and Maven & Primary; Holm $p$ = 1 & No demonstrated difference \\
\midrule
Direction under other campaign economics & Exploratory; \nScenarios{} combinations; unadjusted & TabPFN-3.5 ahead on
the real-operator datasets within the tested grid; size varies \\
GBDTs tuned on EMP instead of AUC & Exploratory; \nSeeds{} splits of five datasets; unadjusted & No gap moves by more
than \maxEmpTuneShift{} or changes sign \\
Larger tuning budget, duplicate-aware split & Exploratory; five splits of three datasets & Conclusions unchanged where
run \\
TabICL v2 calibration on Orange & Exploratory; CITL and policy regret & Recalibrate before thresholding \\
E-Commerce replication & Exploratory; five splits & Tuning narrows the earlier gap without closing it \\
\bottomrule
\end{tabular}
\end{table}

For an operator, the results suggest an order of operations rather than a winner.
\begin{itemize}
\item \textbf{Tune the GBDT first.} A default GBDT overstates what a \tfm adds; tuning removed about a third of the
  earlier headline gap (Section~\ref{sec:rq5}).
\item \textbf{Compare candidates on held-out campaign profit, on one's own customers.} Our gains are specific to
  datasets and model generations, and the significant ones come from two operators. They do not transfer to another
  operator's customers without a check on that operator's data.
\item \textbf{Consider TabPFN-3.5 when its licence terms are acceptable} and it leads on such a check. Its raw
  probabilities were the most consistently close to the observed churn rate among the \tfms we tested, though an
  operator should still check calibration on its own data, and with a cached context its scoring cost is small
  (Section~\ref{sec:rq6}).
\item \textbf{Keep the tuned GBDT} when the \tfm's lead is not significant on one's own data, when only openly licensed
  models are acceptable and TabICL v2 does not lead on one's own data.
\item \textbf{Recalibrate TabICL v2 before thresholding.} Its ranking is competitive, but its raw probabilities
  underpredicted churn on Orange and Cell2Cell, and a fixed threshold then misses churners worth contacting
  (Section~\ref{sec:rq3}).
\item \textbf{Do not balance the context when the training set fits.} It reduced the observed advantage and distorted
  the probabilities (Section~\ref{sec:rq4}).
\end{itemize}
The licence can outweigh a measured advantage: an operator cannot use the outputs of the TabPFN weights we evaluated
for commercial decisions without a separate agreement, however large the gain.

The finding most likely to change with more evidence is the Orange gain. It is consistent across splits but not
significant after correction, the splits resample a single operator's customers, and a feature-engineered GBDT
\cite{tschalzev2026tabprep} could narrow it. The finding least likely to change is the calibration pattern, which is
consistent across splits and matches the underprediction reported for credit risk \cite{baesens2026creditrisk}.

\section{Limitations}\label{sec:lim}

Each limitation below restricts a specific conclusion. We list them in order of how much they restrict the paper's
main claim.

\begin{enumerate}
\item \textbf{Three real-operator datasets.} Only Cell2Cell, Iranian and Orange come from operators, and the two
  significant TabPFN-3.5 results come from two of them. Because of this, we cannot claim that TabPFN-3.5's gains
  generalise to telecom operators in general, to other industries, or to operators with millions of customers, where
  the context would no longer hold the full training set.
\item \textbf{Repeated splits are not independent operators.} The \nSeeds{} splits resample the same customers. They
  measure how stable a comparison is within a dataset, not how it varies across operators, so the count of
  significant comparisons is not an estimate of how many operators would gain.
\item \textbf{Limited statistical power.} With \nSeeds{} splits and a correction over \nPrimary{} comparisons, moderate
  differences can go undetected. A non-significant result, such as Orange, therefore does not show that the models
  are equivalent, and more splits or datasets could move it either way.
\item \textbf{Assumed campaign economics.} EMP and the threshold policy use assumed values of customer value, incentive
  cost and acceptance rate. The grid in Section~\ref{sec:scn} shows how conclusions move with these values, but no
  result here is profit measured in a real campaign, and e-Profits, which uses customer-level value, is available for
  three datasets only.
\item \textbf{Primary baselines tuned on AUC.} The confirmatory gaps are measured against GBDTs tuned on AUC. Retuning
  on EMP did not produce a stronger baseline (Section~\ref{sec:sens}), but that check is exploratory, uses the same
  small budgets, scores trials by EMP on a small inner split while early stopping still uses AUC, and does not cover
  other profit-aware objectives such as cost-sensitive losses. We therefore cannot claim that no profit-aware tuning
  of the GBDTs would narrow the gaps.
\item \textbf{No feature engineering.} Feature engineering can change model comparisons \cite{tschalzev2026tabprep};
  our conclusions hold for the features as distributed.
\item \textbf{Reduced scope of the robustness checks.} The tuning-budget, duplicate and context checks, and the
  replication, use five splits and a subset of datasets; XGBoost was not rerun with 200 trials. They can show that a
  conclusion did not change where they were run, not that it would not change elsewhere.
\item \textbf{Versions, settings and hardware.} We used one GPU type, the default inference settings of each \tfm, and
  specific model versions, checkpoints and licence texts (Appendix~\ref{app:repro}). Any of these can change. We did
  not evaluate the API-only TabPFN-3.5-Plus and -Thinking variants or the other \tfms covered by
  \cite{seyedzadeh2026churn}.
\end{enumerate}

\section{Conclusion}

We asked whether tabular foundation models select more profitable retention targets than tuned gradient-boosted trees.
On three real-operator and two fictional telecom datasets, the answer is selective. TabPFN-3.5 earns significantly more
expected campaign profit than each of three tuned GBDTs on Cell2Cell and Iranian. On Orange it is ahead in every
split, but that gain is not significant after correction, and on the two fictional datasets no model can be told
apart. Among the \tfms, its raw probabilities and TabPFN-3.5-Fast's were the most consistently close to the observed churn
rate, whereas TabICL v2 needed recalibration before its probabilities could drive a campaign. These results describe the datasets
studied, under assumed campaign economics; they do not establish an advantage for telecom operators in general. The
TabPFN-3.5 weights we used cannot be used in production under their licence (TabPFN-3.5 License v1.0, read in October
2026) without a commercial agreement. We release the pipeline and every per-split prediction so that these conclusions
can be re-checked as models, datasets and licences change.

\section*{Data and code availability}\label{sec:availability}
\begin{sloppypar}
The version of the code and results described here is archived on Zenodo as release v1.7
(\url{https://doi.org/10.5281/zenodo.24225904}) and maintained at
\url{https://github.com/ihpolash/tfm-churn-profit-audit}. The repository contains the prediction and analysis code as
scripts and as notebooks, the Kaggle kernel definitions that ran them, the per-split predictions and split indices of
all \nRuns{} runs, the hyperparameters selected for each of the \nHyperConfigs{} tuned GBDT fits
(\texttt{results/gbdt\_hyperparameters.csv}), the analysis outputs, and the generator and audit for every table and number in this paper. The
datasets are public (Table~\ref{tab:data}) but are not redistributed, because several are licensed for non-commercial
or attributed use only. The labels needed to recompute metrics from the predictions are regenerated from the original
sources by the released code.
\end{sloppypar}

\section*{Reproducibility}
Every model run writes its predictions, split indices, timing and package versions; the analysis recomputes every
metric from those files. Every table, figure and in-text number in this paper is generated by
\texttt{paper/make\_tables.py} from the released analysis outputs, and \texttt{paper/audit.py} re-derives them
without rebuilding the pipeline. The audit first checks that regeneration reproduces the committed tables exactly. It then recomputes every corrected
$t$-test, Holm adjustment and intersection--union $p$-value from the per-split differences, and confirms that each
reference GBDT is the best on the calibration split. It verifies that all runs succeeded with identical and disjoint
splits, that the primary family holds exactly the \nPrimary{} stated comparisons, and that relative gains use the stated
denominator. It checks that the released hyperparameters cover every dataset, split and tuned GBDT, and that each AUC-tuned
configuration reproduces the released predictions. Finally, it checks that the robustness checks and the
campaign-economics grid cover what the paper states,
and that the package versions logged by every run match Appendix~\ref{app:repro}. It needs only the Python standard library.

The original runs did not log the GBDT configurations Optuna selected. We recovered them by rerunning the tuning with
the same seeds and splits and with logging; all \nHyperReproduce{} of \nHyperAUC{} AUC-tuned reruns reproduce the released
predictions exactly, so the released configurations are the ones behind the reported results.

The release supports three levels of reproduction, from cheapest to complete. (1)~\texttt{python3 paper/audit.py}
regenerates every table and number from the released analysis outputs in seconds. (2)~The analysis notebook recomputes
those outputs from the released per-split predictions, given the labels, which the prediction code regenerates from the
public sources. (3)~The prediction notebook reruns every model on Kaggle, about 10 to 15 T4-hours in total, and needs a
TabPFN licence token. Model versions, checkpoints and package versions can change; Appendix~\ref{app:repro} lists the
ones used, and exact reproduction requires them.

\section*{Declaration on the use of generative AI}
The author used an AI assistant (Claude, Anthropic) for implementation and execution of the experiment code, for
literature search and reference checking, for table and figure generation, and for drafting and editing this
manuscript. The research question was chosen by the author; the benchmark design, evaluation protocol and
interpretation of the results were developed in iterative discussion with the assistant, and every design decision and
conclusion was reviewed and approved by the author. All numerical results were produced by the released code. Every
table, figure and in-text number is generated by a released script from the released analysis outputs, which are
computed from the released per-split predictions, and an audit notebook re-derives them before submission. Every
reference was checked against its primary source or an authoritative index. The author takes full responsibility for
all content of this paper, irrespective of how it was drafted.

\bibliographystyle{plainnat}
\bibliography{references}

\appendix
\section{Reproducibility Details}\label{app:repro}

\begin{table}[h]
\caption{Software, hardware and baseline search spaces. GBDTs: Optuna TPE, validation AUC on an inner 80/20 split
(EMP in the tuning-objective check of Section~\ref{sec:sens}), early stopping on validation AUC after 100 rounds (up to
3,000 trees), refit on the full training split. Selected configurations: \texttt{results/gbdt\_hyperparameters.csv}.}
\label{tab:repro}
\centering\scriptsize
\begin{tabular}{@{}p{0.2\textwidth}p{0.74\textwidth}@{}}
\toprule
Software & tabpfn 9.2.0 (TabPFN-2.5, 3.5, 3.5-Fast); tabicl 2.2.0 (checkpoint 2026-02-12); lightgbm 4.6.0; xgboost 3.4.1;
catboost 1.2.10; interpret 0.7.8; optuna 5.0.0; scikit-learn 1.6.1; torch 2.11.0 (CUDA 12.8) \\
Hardware & Kaggle: NVIDIA T4 (16\,GB) for \tfms; CPU sessions for all other models \\
LightGBM & 40 trials (200 in Section~\ref{sec:sens}); learning rate $[0.005, 0.1]$ log; leaves $[8, 128]$ log; min child
samples $[5, 200]$ log; subsample $[0.5, 1]$; column sample $[0.3, 1]$; L2 $[10^{-3}, 10]$ log \\
XGBoost & 40 trials; learning rate $[0.005, 0.1]$ log; depth $[3, 10]$; min child weight $[0.5, 50]$ log; subsample
$[0.5, 1]$; column sample $[0.3, 1]$; L2 $[10^{-3}, 10]$ log \\
CatBoost & 25 trials (200 in Section~\ref{sec:sens}); learning rate $[0.01, 0.2]$ log; depth $[4, 10]$; L2 $[1, 30]$
log; random strength $[0.1, 10]$ log; bagging temperature $[0, 1]$ \\
EBM, LR & defaults; LR with one-hot encoding (up to 30 categories), median imputation with missing indicators, scaling \\
\bottomrule
\end{tabular}
\end{table}

\section{Additional Results}

\begin{table}[h]
\caption{Exploratory: PR-AUC difference in points ($\times100$) and EMP difference relative to the best tuned GBDT,
over \nSeeds{} splits. $\ast$: significant after Holm correction (EMP, primary family only); $\dagger$: two-level
bootstrap interval excludes zero. Reference shown as PR-AUC / EMP (LGBM: LightGBM, XGB: XGBoost, CatB: CatBoost).}
\label{tab:boot}
\centering\scriptsize\setlength{\tabcolsep}{3pt}
\resizebox{\textwidth}{!}{%
\begin{tabular}{@{}lcccccccccc@{}}
\toprule
& \multicolumn{2}{c}{Telco} & \multicolumn{2}{c}{Maven} & \multicolumn{2}{c}{Cell2Cell} & \multicolumn{2}{c}{Iranian} & \multicolumn{2}{c}{Orange} \\
\cmidrule(lr){2-3}\cmidrule(lr){4-5}\cmidrule(lr){6-7}\cmidrule(lr){8-9}\cmidrule(l){10-11}
& \input{tables/bootstrap_refs.tex} 
\tfm & $\Delta$PR & $\Delta$EMP & $\Delta$PR & $\Delta$EMP & $\Delta$PR & $\Delta$EMP & $\Delta$PR & $\Delta$EMP & $\Delta$PR & $\Delta$EMP \\
\midrule
\input{tables/bootstrap.tex} 
\bottomrule
\end{tabular}}
\end{table}

\begin{table}[h]
\caption{Mean PR-AUC and EMP (profit per customer, CLV units) over \nSeeds{} splits; best per column in bold.}
\label{tab:main}
\centering\scriptsize\setlength{\tabcolsep}{3.2pt}
\resizebox{\textwidth}{!}{%
\begin{tabular}{@{}lcccccccccc@{}}
\toprule
& \multicolumn{2}{c}{Telco} & \multicolumn{2}{c}{Maven} & \multicolumn{2}{c}{Cell2Cell} & \multicolumn{2}{c}{Iranian} & \multicolumn{2}{c}{Orange} \\
\cmidrule(lr){2-3}\cmidrule(lr){4-5}\cmidrule(lr){6-7}\cmidrule(lr){8-9}\cmidrule(l){10-11}
Model & PR-AUC & EMP & PR-AUC & EMP & PR-AUC & EMP & PR-AUC & EMP & PR-AUC & EMP \\
\midrule
\input{tables/main_results.tex} 
\bottomrule
\end{tabular}}
\end{table}

\begin{table}[h]
\caption{Mean ROC-AUC over \nSeeds{} splits; best per column in bold.}
\label{tab:auc}
\centering\small
\begin{tabular}{@{}lccccc@{}}
\toprule
Model & Telco & Maven & Cell2Cell & Iranian & Orange \\
\midrule
\input{tables/auc_results.tex} 
\bottomrule
\end{tabular}
\end{table}

\end{document}

%% file: tables/numbers.tex
\newcommand{\anyTunedPFNThreeMavenP}{0.036}

\newcommand{\citlICLCellTwoCellAbs}{1.8}

\newcommand{\citlICLOrangeAbs}{2.1}

\newcommand{\citlPFNTwoCellTwoCellAbs}{2.0}

\newcommand{\conflictGroupsIranian}{14}
\newcommand{\conflictGroupsTelco}{18}

\newcommand{\conflictRowsIranian}{64}
\newcommand{\conflictRowsTelco}{42}
\newcommand{\ctxCellTwoCell}{30,628}

\newcommand{\ctxCitlOrangeAbs}{35.3}
\newcommand{\ctxOrange}{30,000}

\newcommand{\ctxOrangePFNBalAbs}{4.5}

\newcommand{\ctxOrangePFNFullAbs}{14.3}

\newcommand{\ctxOrangePFNHalfAbs}{8.4}
\newcommand{\ctxOrangePFNpBal}{0.541}

\newcommand{\ctxOrangePFNpHalf}{0.201}
\newcommand{\ctxTelco}{4,201}

\newcommand{\dupECommerce}{556}
\newcommand{\dupIranian}{314}

\newcommand{\dupPctECommerce}{9.9}
\newcommand{\dupPctIranian}{10.0}

\newcommand{\dupPctTelco}{0.6}
\newcommand{\dupTelco}{40}

\newcommand{\empAbsPFNThreeCellTwoCellAbs}{0.175}

\newcommand{\empAbsPFNThreeIranianAbs}{0.597}

\newcommand{\empAbsPFNThreeOrangeAbs}{0.082}

\newcommand{\empICLOrange}{-4.5}

\newcommand{\empPFNThreeCellTwoCellAbs}{1.9}

\newcommand{\empPFNThreeIranianAbs}{7.6}

\newcommand{\empPFNThreeMavenAbs}{1.0}

\newcommand{\empPFNThreeOrangeAbs}{14.3}

\newcommand{\empPFNThreeTelcoAbs}{1.0}

\newcommand{\empRefDatasets}{Telco, Maven, Cell2Cell}

\newcommand{\empTunedPFNThreeIranianAbs}{8.5}

\newcommand{\empTunedPFNThreeOrangeAbs}{15.7}

\newcommand{\holmPFNFastCellTwoCell}{0.036}

\newcommand{\holmPFNThreeCellTwoCell}{$<$0.001}
\newcommand{\holmPFNThreeIranian}{0.003}

\newcommand{\holmPFNThreeOrange}{0.155}

\newcommand{\hoursPFNCacheMillion}{1.25}

\newcommand{\iuHolmPFNFastCellTwoCell}{0.046}

\newcommand{\iuHolmPFNThreeCellTwoCell}{$<$0.001}
\newcommand{\iuHolmPFNThreeIranian}{0.043}

\newcommand{\iuSigCases}{TabPFN-3.5 on Cell2Cell; TabPFN-3.5 on Iranian; TabPFN-3.5-Fast on Cell2Cell}
\newcommand{\maxAbsCitlPFNFast}{0.3}
\newcommand{\maxAbsCitlPFNThree}{0.2}
\newcommand{\maxChurnShift}{0.3}
\newcommand{\maxDupShift}{1.6}
\newcommand{\maxEmpTuneShift}{0.014}
\newcommand{\maxRawRegretNonOrange}{3.1}
\newcommand{\maxRegretShiftNonOrange}{0.9}

\newcommand{\nBrierBestPFNThree}{5}

\newcommand{\nECEBestPFNThree}{3}

\newcommand{\nEmpRefDatasets}{3}
\newcommand{\nEmpTuneSignFlip}{0}

\newcommand{\nHolmSig}{6}
\newcommand{\nHolmSigTest}{6}
\newcommand{\nHyperAUC}{165}
\newcommand{\nHyperConfigs}{330}
\newcommand{\nHyperReproduce}{165}
\newcommand{\nIUSig}{3}

\newcommand{\nPrimary}{20}

\newcommand{\nRuns}{850}
\newcommand{\nScenarios}{27}
\newcommand{\nScnSignFlip}{8}
\newcommand{\nSeeds}{10}

\newcommand{\nbpPFNThreeOrange}{0.011}

\newcommand{\perMillionPFNThreeCellTwoCell}{175,000}

\newcommand{\rateOrange}{7.3}

\newcommand{\refEMPPFNThreeOrange}{0.576}

\newcommand{\regICLOrangeIso}{7.5}
\newcommand{\regICLOrangePlatt}{6.3}
\newcommand{\regICLOrangeRaw}{15.7}
\newcommand{\replDefaultPR}{0.906}

\newcommand{\replICLDefaultAbs}{6.1}

\newcommand{\replICLEMPAbs}{4.8}
\newcommand{\replICLSeeds}{5}

\newcommand{\replICLTunedAbs}{4.0}
\newcommand{\replICLTunedP}{0.023}

\newcommand{\replPFNThreeEMPAbs}{8.3}

\newcommand{\replPFNThreeTunedAbs}{5.8}

\newcommand{\replTuned}{CatBoost}
\newcommand{\replTunedPR}{0.927}
\newcommand{\scnAbsMaxCellTwoCell}{0.025}

\newcommand{\scnAbsMaxOrange}{0.059}

\newcommand{\scnPosCellTwoCell}{27}
\newcommand{\scnPosIranian}{27}

\newcommand{\scnPosOrange}{27}

\newcommand{\scnRefMaxOrange}{0.074}

\newcommand{\scnRelMaxCellTwoCellAbs}{65.8}

\newcommand{\scnRelMaxOrangeAbs}{80.7}

\newcommand{\scnSigOrange}{25}

\newcommand{\scnTopSameTelco}{15}

\newcommand{\seedsPosPFNThreeOrange}{10}

\newcommand{\strongDeltaCellTwoCell}{0.0}
\newcommand{\strongDeltaIranian}{-0.8}

\newcommand{\strongDeltaOrange}{+0.1}

\newcommand{\strongPFNThreeCellTwoCellAbs}{1.9}

\newcommand{\strongPFNThreeOrangeAbs}{20.1}

\newcommand{\tauEPthrMaven}{0.04}

\newcommand{\timeICLCacheOrange}{17}
\newcommand{\timeLGBMOrange}{0.8}
\newcommand{\timePFNCacheOrange}{45}
\newcommand{\timePFNUncachedSmallOrange}{1,686}
\newcommand{\topEMPCellTwoCell}{0}
\newcommand{\topEMPIranian}{6}
\newcommand{\topEMPMaven}{7}
\newcommand{\topEMPOrange}{1}
\newcommand{\topEMPTelco}{7}

%% file: tables/datasets.tex
Telco \cite{telco} & IBM sample data (fictional) & 7,003 & 19 & 26.4 & 40 & yes & main \\
Maven \cite{maven} & Maven Analytics (fictional) & 6,589 & 34 & 28.4 & 0 & yes & main \\
Cell2Cell \cite{cell2cell} & US wireless carrier & 51,047 & 56 & 28.8 & 0 & yes & main \\
Iranian \cite{iranian} & Iranian telecom operator & 2,836 & 15 & 15.6 & 314 & no & main \\
Orange \cite{guyon2009kddcup} & Orange, KDD Cup 2009 & 50,000 & 202 & 7.3 & 0 & no & main \\
E-Commerce \cite{ecommerce} & Kaggle, origin unstated & 5,074 & 18 & 16.6 & 556 & no & replication \\

%% file: tables/primary.tex
\multirow{4}{*}{Telco} & \multirow{4}{*}{\shortstack[l]{LightGBM\\10.232}} & TabICL v2 & $+0.066$ & $+0.6$ & $[-0.024, +0.156]$ & 9/10 & 0.133 & 1.000 & 1.000 \\
 & & TabPFN-2.5 & $+0.060$ & $+0.6$ & $[-0.048, +0.168]$ & 8/10 & 0.240 & 1.000 & 1.000 \\
 & & TabPFN-3.5 & $+0.104$ & $+1.0$ & $[-0.014, +0.222]$ & 9/10 & 0.078 & 1.000 & 1.000 \\
 & & TabPFN-3.5-Fast & $+0.090$ & $+0.9$ & $[-0.020, +0.200]$ & 9/10 & 0.097 & 1.000 & 1.000 \\
\midrule
\multirow{4}{*}{Maven} & \multirow{4}{*}{\shortstack[l]{XGBoost\\13.059}} & TabICL v2 & $+0.042$ & $+0.3$ & $[-0.129, +0.213]$ & 6/10 & 0.589 & 1.000 & 1.000 \\
 & & TabPFN-2.5 & $+0.090$ & $+0.7$ & $[-0.083, +0.264]$ & 8/10 & 0.270 & 1.000 & 1.000 \\
 & & TabPFN-3.5 & $+0.135$ & $+1.0$ & $[-0.041, +0.312]$ & 8/10 & 0.116 & 1.000 & 1.000 \\
 & & TabPFN-3.5-Fast & $+0.095$ & $+0.7$ & $[-0.070, +0.260]$ & 8/10 & 0.225 & 1.000 & 1.000 \\
\midrule
\multirow{4}{*}{Cell2Cell} & \multirow{4}{*}{\shortstack[l]{LightGBM\\9.176}} & TabICL v2 & $+0.042$ & $+0.5$ & $[-0.007, +0.091]$ & 9/10 & 0.086 & 1.000 & 1.000 \\
 & & TabPFN-2.5 & $+0.032$ & $+0.4$ & $[-0.012, +0.077]$ & 8/10 & 0.136 & 1.000 & 1.000 \\
 & & TabPFN-3.5 & $+0.175$ & $+1.9$ & $[+0.133, +0.216]$ & 10/10 & $<$0.001 & \textbf{$<$0.001} & \textbf{$<$0.001} \\
 & & TabPFN-3.5-Fast & $+0.104$ & $+1.1$ & $[+0.048, +0.161]$ & 10/10 & 0.002 & \textbf{0.036} & \textbf{0.046} \\
\midrule
\multirow{4}{*}{Iranian} & \multirow{4}{*}{\shortstack[l]{LightGBM\\7.860}} & TabICL v2 & $+0.584$ & $+7.4$ & $[+0.333, +0.835]$ & 10/10 & $<$0.001 & \textbf{0.009} & 0.051 \\
 & & TabPFN-2.5 & $+0.543$ & $+6.9$ & $[+0.302, +0.785]$ & 10/10 & $<$0.001 & \textbf{0.011} & 0.093 \\
 & & TabPFN-3.5 & $+0.597$ & $+7.6$ & $[+0.382, +0.813]$ & 10/10 & $<$0.001 & \textbf{0.003} & \textbf{0.043} \\
 & & TabPFN-3.5-Fast & $+0.561$ & $+7.1$ & $[+0.317, +0.806]$ & 10/10 & $<$0.001 & \textbf{0.010} & 0.082 \\
\midrule
\multirow{4}{*}{Orange} & \multirow{4}{*}{\shortstack[l]{LightGBM\\0.576}} & TabICL v2 & $-0.026$ & $-4.5$ & $[-0.101, +0.049]$ & 4/10 & 0.457 & 1.000 & 1.000 \\
 & & TabPFN-2.5 & $-0.017$ & $-3.0$ & $[-0.070, +0.035]$ & 2/10 & 0.474 & 1.000 & 1.000 \\
 & & TabPFN-3.5 & $+0.082$ & $+14.3$ & $[+0.024, +0.140]$ & 10/10 & 0.011 & 0.155 & 0.155 \\
 & & TabPFN-3.5-Fast & $+0.045$ & $+7.8$ & $[-0.011, +0.101]$ & 8/10 & 0.102 & 1.000 & 1.000 \\

%% file: tables/rank_agreement.tex
Telco & $0.52$ & 7/10 & $0.23$ & 10/10 & $0.28$ & 10/10 \\
Maven & $0.67$ & 7/10 & $0.47$ & 5/10 & $0.04$ & 7/10 \\
Cell2Cell & $0.88$ & 0/10 & $0.42$ & 1/10 & $0.41$ & 1/10 \\
Iranian & $0.79$ & 6/10 & -- & -- & -- & -- \\
Orange & $0.83$ & 1/10 & -- & -- & -- & -- \\

%% file: tables/calibration.tex
\multirow{5}{*}{Telco} & LightGBM & $-0.1$ & 0.1414 & 0.051 & 0.029 & 1.8 & 1.3 & 1.4 \\
 & TabICL v2 & $+0.7$ & 0.1338 & 0.030 & 0.028 & 1.5 & 1.5 & 1.1 \\
 & TabPFN-2.5 & $+0.6$ & 0.1340 & 0.028 & 0.031 & 1.4 & 1.6 & 1.4 \\
 & TabPFN-3.5 & $0.0$ & 0.1333 & 0.031 & 0.029 & 1.3 & 1.7 & 1.2 \\
 & TabPFN-3.5-Fast & $0.0$ & 0.1333 & 0.029 & 0.030 & 1.5 & 1.5 & 1.2 \\
\midrule
\multirow{5}{*}{Maven} & XGBoost & $+0.6$ & 0.0830 & 0.027 & 0.028 & 0.9 & 1.2 & 0.9 \\
 & TabICL v2 & $+0.8$ & 0.0809 & 0.026 & 0.024 & 1.1 & 1.1 & 0.9 \\
 & TabPFN-2.5 & $+0.7$ & 0.0812 & 0.022 & 0.026 & 0.9 & 1.0 & 1.1 \\
 & TabPFN-3.5 & $+0.2$ & 0.0769 & 0.020 & 0.023 & 0.7 & 1.0 & 0.9 \\
 & TabPFN-3.5-Fast & $+0.2$ & 0.0779 & 0.024 & 0.023 & 0.8 & 1.0 & 0.9 \\
\midrule
\multirow{5}{*}{Cell2Cell} & LightGBM & $+0.1$ & 0.1866 & 0.011 & 0.011 & 0.5 & 0.5 & 0.4 \\
 & TabICL v2 & $+1.8$ & 0.1863 & 0.019 & 0.012 & 0.6 & 0.5 & 0.5 \\
 & TabPFN-2.5 & $+2.0$ & 0.1862 & 0.021 & 0.012 & 0.7 & 0.4 & 0.5 \\
 & TabPFN-3.5 & $-0.2$ & 0.1826 & 0.012 & 0.014 & 0.5 & 0.5 & 0.6 \\
 & TabPFN-3.5-Fast & $-0.3$ & 0.1839 & 0.012 & 0.013 & 0.4 & 0.4 & 0.4 \\
\midrule
\multirow{5}{*}{Iranian} & LightGBM & $+0.8$ & 0.0313 & 0.033 & 0.024 & 2.3 & 2.6 & 2.1 \\
 & TabICL v2 & $-0.1$ & 0.0118 & 0.014 & 0.011 & 0.8 & 1.0 & 1.0 \\
 & TabPFN-2.5 & $0.0$ & 0.0139 & 0.016 & 0.011 & 1.0 & 0.9 & 0.8 \\
 & TabPFN-3.5 & $-0.1$ & 0.0106 & 0.013 & 0.012 & 0.9 & 0.9 & 0.9 \\
 & TabPFN-3.5-Fast & $-0.2$ & 0.0116 & 0.014 & 0.011 & 0.9 & 0.7 & 0.9 \\
\midrule
\multirow{5}{*}{Orange} & LightGBM & $0.0$ & 0.0637 & 0.008 & 0.005 & 8.2 & 7.1 & 5.8 \\
 & TabICL v2 & $+2.1$ & 0.0642 & 0.023 & 0.004 & 15.7 & 7.5 & 6.3 \\
 & TabPFN-2.5 & $+0.2$ & 0.0637 & 0.008 & 0.006 & 7.7 & 6.1 & 6.0 \\
 & TabPFN-3.5 & $0.0$ & 0.0624 & 0.005 & 0.005 & 5.5 & 5.7 & 4.9 \\
 & TabPFN-3.5-Fast & $0.0$ & 0.0628 & 0.006 & 0.005 & 5.4 & 8.2 & 5.2 \\

%% file: tables/context.tex
Orange & TabPFN-3.5 & full (natural rate) & 0.748 & 0.238 & 0.658 & $+14.3\%$ & 10/10 & 0.011 & $0.0$ \\
Orange & TabPFN-3.5 & 1:2 & 0.739 & 0.219 & 0.611 & $+8.4\%$ & 5/5 & 0.201 & $-21.6$ \\
Orange & TabPFN-3.5 & 1:1 & 0.736 & 0.208 & 0.589 & $+4.5\%$ & 3/5 & 0.541 & $-35.3$ \\
\midrule
Orange & TabICL v2 & full (natural rate) & 0.727 & 0.210 & 0.550 & $-4.5\%$ & 4/10 & 0.457 & $+2.1$ \\
Orange & TabICL v2 & 1:2 & 0.730 & 0.209 & 0.557 & $-1.2\%$ & 3/5 & 0.875 & $-19.9$ \\
Orange & TabICL v2 & 1:1 & 0.727 & 0.196 & 0.525 & $-6.9\%$ & 0/5 & 0.309 & $-35.1$ \\
\midrule
Cell2Cell & TabPFN-3.5 & full (natural rate) & 0.702 & 0.495 & 9.350 & $+1.9\%$ & 10/10 & $<$0.001 & $-0.2$ \\
Cell2Cell & TabPFN-3.5 & 1:2 & 0.704 & 0.498 & 9.357 & $+1.9\%$ & 5/5 & 0.006 & $-4.4$ \\
Cell2Cell & TabPFN-3.5 & 1:1 & 0.700 & 0.491 & 9.324 & $+1.5\%$ & 5/5 & 0.021 & $-19.1$ \\
\midrule
Cell2Cell & TabICL v2 & full (natural rate) & 0.687 & 0.472 & 9.218 & $+0.5\%$ & 9/10 & 0.086 & $+1.8$ \\
Cell2Cell & TabICL v2 & 1:2 & 0.689 & 0.474 & 9.226 & $+0.4\%$ & 4/5 & 0.301 & $-2.7$ \\
Cell2Cell & TabICL v2 & 1:1 & 0.686 & 0.468 & 9.197 & $+0.1\%$ & 3/5 & 0.750 & $-19.1$ \\

%% file: tables/emptuned.tex
\multirow{4}{*}{Telco} & TabICL v2 & $+0.066$ & $+0.6\%$ & \multirow{4}{*}{XGBoost (EMP)} & $+0.080$ & $+0.8\%$ & 9/10 & 0.164 \\
 & TabPFN-2.5 & $+0.060$ & $+0.6\%$ &  & $+0.074$ & $+0.7\%$ & 7/10 & 0.326 \\
 & TabPFN-3.5 & $+0.104$ & $+1.0\%$ &  & $+0.118$ & $+1.2\%$ & 10/10 & 0.083 \\
 & TabPFN-3.5-Fast & $+0.090$ & $+0.9\%$ &  & $+0.104$ & $+1.0\%$ & 10/10 & 0.133 \\
\midrule
\multirow{4}{*}{Maven} & TabICL v2 & $+0.042$ & $+0.3\%$ & \multirow{4}{*}{CatBoost (EMP)} & $+0.030$ & $+0.2\%$ & 6/10 & 0.578 \\
 & TabPFN-2.5 & $+0.090$ & $+0.7\%$ &  & $+0.078$ & $+0.6\%$ & 9/10 & 0.116 \\
 & TabPFN-3.5 & $+0.135$ & $+1.0\%$ &  & $+0.123$ & $+0.9\%$ & 9/10 & 0.036 \\
 & TabPFN-3.5-Fast & $+0.095$ & $+0.7\%$ &  & $+0.082$ & $+0.6\%$ & 8/10 & 0.101 \\
\midrule
\multirow{4}{*}{Cell2Cell} & TabICL v2 & $+0.042$ & $+0.5\%$ & \multirow{4}{*}{XGBoost (EMP)} & $+0.043$ & $+0.5\%$ & 10/10 & 0.094 \\
 & TabPFN-2.5 & $+0.032$ & $+0.4\%$ &  & $+0.034$ & $+0.4\%$ & 8/10 & 0.213 \\
 & TabPFN-3.5 & $+0.175$ & $+1.9\%$ &  & $+0.176$ & $+1.9\%$ & 10/10 & $<$0.001 \\
 & TabPFN-3.5-Fast & $+0.104$ & $+1.1\%$ &  & $+0.106$ & $+1.2\%$ & 10/10 & 0.001 \\
\midrule
\multirow{4}{*}{Iranian} & TabICL v2 & $+0.584$ & $+7.4\%$ & \multirow{4}{*}{LightGBM} & $+0.584$ & $+7.4\%$ & 10/10 & $<$0.001 \\
 & TabPFN-2.5 & $+0.543$ & $+6.9\%$ &  & $+0.543$ & $+6.9\%$ & 10/10 & $<$0.001 \\
 & TabPFN-3.5 & $+0.597$ & $+7.6\%$ &  & $+0.597$ & $+7.6\%$ & 10/10 & $<$0.001 \\
 & TabPFN-3.5-Fast & $+0.561$ & $+7.1\%$ &  & $+0.561$ & $+7.1\%$ & 10/10 & $<$0.001 \\
\midrule
\multirow{4}{*}{Orange} & TabICL v2 & $-0.026$ & $-4.5\%$ & \multirow{4}{*}{LightGBM} & $-0.026$ & $-4.5\%$ & 4/10 & 0.457 \\
 & TabPFN-2.5 & $-0.017$ & $-3.0\%$ &  & $-0.017$ & $-3.0\%$ & 2/10 & 0.474 \\
 & TabPFN-3.5 & $+0.082$ & $+14.3\%$ &  & $+0.082$ & $+14.3\%$ & 10/10 & 0.011 \\
 & TabPFN-3.5-Fast & $+0.045$ & $+7.8\%$ &  & $+0.045$ & $+7.8\%$ & 8/10 & 0.102 \\

%% file: tables/strong.tex
Cell2Cell & TabPFN-3.5 & LightGBM & $+1.9\%$ & LightGBM (200 trials) & $+1.9\%$ & 5/5 & 0.005 \\
Cell2Cell & TabICL v2 & LightGBM & $+0.4\%$ & LightGBM (200 trials) & $+0.4\%$ & 4/5 & 0.312 \\
Iranian & TabPFN-3.5 & LightGBM & $+7.9\%$ & LightGBM & $+7.9\%$ & 5/5 & 0.011 \\
Iranian & TabICL v2 & LightGBM & $+7.5\%$ & LightGBM & $+7.5\%$ & 5/5 & 0.020 \\
Orange & TabPFN-3.5 & CatBoost & $+20.1\%$ & CatBoost & $+20.1\%$ & 5/5 & 0.005 \\
Orange & TabICL v2 & CatBoost & $-0.2\%$ & CatBoost & $-0.2\%$ & 3/5 & 0.972 \\

%% file: tables/dupgroup.tex
Telco & 40 & 18 (42) & 26.5 / 26.4 & TabPFN-3.5 & $+1.0\%$ & $+0.8\%$ & 0.288 \\
Telco & 40 & 18 (42) & 26.5 / 26.4 & TabICL v2 & $+0.6\%$ & $+0.7\%$ & 0.322 \\
Iranian & 314 & 14 (64) & 15.7 / 15.6 & TabPFN-3.5 & $+7.6\%$ & $+7.5\%$ & 0.001 \\
Iranian & 314 & 14 (64) & 15.7 / 15.6 & TabICL v2 & $+7.4\%$ & $+6.9\%$ & 0.005 \\
E-Commerce & 556 & 0 (0) & 16.8 / 16.6 & TabPFN-3.5 & $+8.3\%$ & $+9.4\%$ & 0.035 \\
E-Commerce & 556 & 0 (0) & 16.8 / 16.6 & TabICL v2 & $+4.8\%$ & $+5.8\%$ & 0.148 \\

%% file: tables/scenarios.tex
Telco & TabPFN-3.5 & 15/27 & $0.0$ & $-0.008$ & 35.554 & $+8.9$ & $+0.022$ & 0.251 & 24/27 & 5/27 \\
Maven & TabPFN-3.5 & 24/27 & $+0.4$ & $+0.260$ & 67.310 & $+5.5$ & $+0.057$ & 1.033 & 27/27 & 6/27 \\
Cell2Cell & TabPFN-3.5 & 27/27 & $0.0$ & $+0.007$ & 66.767 & $+65.8$ & $+0.025$ & 0.038 & 27/27 & 23/27 \\
Iranian & TabPFN-3.5 & 27/27 & $+2.0$ & $+0.727$ & 37.158 & $+29.8$ & $+0.220$ & 0.741 & 27/27 & 27/27 \\
Orange & TabPFN-3.5 & 27/27 & $+0.5$ & $+0.060$ & 13.059 & $+80.7$ & $+0.059$ & 0.074 & 27/27 & 25/27 \\

%% file: tables/replication.tex
XGBoost (library default) & 0.906 (0.019) & 8.090 & -- & -- & -- \\
XGBoost & 0.923 (0.019) & 8.119 & -- & -- & -- \\
LightGBM & 0.922 (0.016) & 8.116 & -- & -- & -- \\
CatBoost & 0.927 (0.017) & 8.119 & -- & -- & -- \\
\midrule
TabICL v2 & 0.967 (0.012) & 8.505 & $\mathbf{+6.1}^{\dagger}$ & $\mathbf{+4.0}^{\dagger}$ & $\mathbf{+4.8\%}^{\dagger}$ \\
TabPFN-2.5 & 0.966 (0.012) & 8.512 & $\mathbf{+6.0}^{\dagger}$ & $\mathbf{+3.9}^{\dagger}$ & $\mathbf{+4.9\%}^{\dagger}$ \\
TabPFN-3.5 & 0.984 (0.008) & 8.790 & $\mathbf{+7.8}^{\dagger}$ & $\mathbf{+5.8}^{\dagger}$ & $\mathbf{+8.3\%}^{\dagger}$ \\

%% file: tables/timing.tex
LightGBM (1,000 trees, CPU) & 0.8 & 0.5 & 0.8 \\
TabICL v2 & 2.5 & 18.8 & 89.2 \\
TabICL v2, KV cache & -- & -- & 16.7 \\
TabPFN-3.5 & 8.8 & 49.4 & 1,686$^\dagger$ \\
TabPFN-3.5, cached context & 8.1 & 30.5 & 44.9 \\
TabPFN-3.5-Fast & 2.7 & 12.4 & 25.9 \\
TabPFN-2.5 & 9.6 & 3,065$^\dagger$ & 10,122$^\dagger$ \\

%% file: tables/bootstrap_refs.tex
\multicolumn{2}{c}{\scriptsize vs.\ XGB / LGBM} & \multicolumn{2}{c}{\scriptsize vs.\ CatB / XGB} & \multicolumn{2}{c}{\scriptsize vs.\ XGB / LGBM} & \multicolumn{2}{c}{\scriptsize vs.\ LGBM / LGBM} & \multicolumn{2}{c}{\scriptsize vs.\ CatB / LGBM} \\

%% file: tables/bootstrap.tex
TabICL v2 & $\mathbf{+0.6}^{\dagger}$ & $\mathbf{+0.6\%}^{\dagger}$ & $+0.3$ & $+0.3\%$ & $\mathbf{+0.5}^{\dagger}$ & $\mathbf{+0.5\%}^{\dagger}$ & $\mathbf{+4.6}^{\dagger}$ & $\mathbf{+7.4\%}^{\ast}$ & $0.0$ & $-4.5\%$ \\
TabPFN-2.5 & $+0.5$ & $\mathbf{+0.6\%}^{\dagger}$ & $+0.2$ & $\mathbf{+0.7\%}^{\dagger}$ & $\mathbf{+0.6}^{\dagger}$ & $\mathbf{+0.4\%}^{\dagger}$ & $\mathbf{+4.4}^{\dagger}$ & $\mathbf{+6.9\%}^{\ast}$ & $-0.5$ & $-3.0\%$ \\
TabPFN-3.5 & $\mathbf{+0.7}^{\dagger}$ & $\mathbf{+1.0\%}^{\dagger}$ & $\mathbf{+1.2}^{\dagger}$ & $\mathbf{+1.0\%}^{\dagger}$ & $\mathbf{+2.9}^{\dagger}$ & $\mathbf{+1.9\%}^{\ast}$ & $\mathbf{+5.0}^{\dagger}$ & $\mathbf{+7.6\%}^{\ast}$ & $\mathbf{+2.8}^{\dagger}$ & $\mathbf{+14.3\%}^{\dagger}$ \\
TabPFN-3.5-Fast & $\mathbf{+0.7}^{\dagger}$ & $\mathbf{+0.9\%}^{\dagger}$ & $\mathbf{+0.9}^{\dagger}$ & $+0.7\%$ & $\mathbf{+2.0}^{\dagger}$ & $\mathbf{+1.1\%}^{\ast}$ & $\mathbf{+4.9}^{\dagger}$ & $\mathbf{+7.1\%}^{\ast}$ & $\mathbf{+1.9}^{\dagger}$ & $\mathbf{+7.8\%}^{\dagger}$ \\

%% file: tables/main_results.tex
Logistic regression & 0.647 & 10.294 & 0.830 & 12.769 & 0.388 & 8.654 & 0.754 & 6.356 & 0.173 & 0.409 \\
EBM & 0.661 & 10.246 & 0.894 & 13.076 & 0.460 & 9.117 & 0.936 & 7.699 & 0.194 & 0.534 \\
LightGBM & 0.657 & 10.232 & 0.890 & 13.067 & 0.468 & 9.176 & 0.941 & 7.860 & 0.206 & 0.576 \\
XGBoost & 0.659 & 10.241 & 0.891 & 13.059 & 0.466 & 9.175 & 0.936 & 7.791 & 0.200 & 0.565 \\
CatBoost & 0.659 & 10.215 & 0.893 & 13.070 & 0.466 & 9.151 & 0.935 & 7.709 & 0.210 & 0.557 \\
\midrule
TabICL v2 & 0.666 & 10.298 & 0.895 & 13.101 & 0.472 & 9.218 & 0.987 & 8.444 & 0.210 & 0.550 \\
TabPFN-2.5 & 0.664 & 10.292 & 0.894 & 13.149 & 0.473 & 9.208 & 0.985 & 8.403 & 0.204 & 0.559 \\
TabPFN-3.5 & \textbf{0.667} & \textbf{10.336} & \textbf{0.904} & \textbf{13.194} & \textbf{0.495} & \textbf{9.350} & \textbf{0.990} & \textbf{8.457} & \textbf{0.238} & \textbf{0.658} \\
TabPFN-3.5-Fast & 0.666 & 10.322 & 0.902 & 13.154 & 0.487 & 9.280 & 0.989 & 8.421 & 0.229 & 0.621 \\

%% file: tables/auc_results.tex
Logistic regression & 0.845 & 0.920 & 0.620 & 0.932 & 0.702 \\
EBM & 0.846 & 0.944 & 0.682 & 0.983 & 0.728 \\
LightGBM & 0.844 & 0.942 & 0.685 & 0.987 & 0.734 \\
XGBoost & 0.846 & 0.942 & 0.685 & 0.985 & 0.733 \\
CatBoost & 0.845 & 0.943 & 0.684 & 0.985 & 0.728 \\
\midrule
TabICL v2 & 0.849 & 0.945 & 0.687 & 0.997 & 0.727 \\
TabPFN-2.5 & 0.848 & 0.946 & 0.688 & 0.996 & 0.733 \\
TabPFN-3.5 & \textbf{0.850} & \textbf{0.950} & \textbf{0.702} & \textbf{0.997} & \textbf{0.748} \\
TabPFN-3.5-Fast & 0.849 & 0.948 & 0.697 & 0.997 & 0.742 \\